\documentclass{article}

\usepackage{arxiv}

\usepackage[utf8]{inputenc} 
\usepackage[T1]{fontenc}    
\usepackage{hyperref}       
\usepackage{url}            
\usepackage{booktabs}       
\usepackage{amsfonts}       
\usepackage{nicefrac}       
\usepackage{microtype}      
\usepackage{lipsum}
\usepackage{graphicx}
\usepackage{amsmath}
\usepackage{float}
\usepackage{subfigure}
\usepackage[tableposition=below]{caption}
\usepackage{esvect}

\usepackage{multirow,makecell}
\usepackage{multicol}

\usepackage[linesnumbered,ruled]{algorithm2e}

\DontPrintSemicolon 

\DeclareMathOperator*{\argmax}{arg\,max}
\let\oldnl\nl
\newcommand{\nonl}{\renewcommand{\nl}{\let\nl\oldnl}}

\title{Feature Acquisition using Monte Carlo Tree Search}

\author{
 Sungsoo Lim \thanks{Corresponding author} \\
  Engineering Sciences and Applied Mathematics \\
  Northwestern University \\
  Evanston, IL 60201 \\
  \texttt{sungsoolim2024@u.northwestern.edu} \\
  \AND
 Diego Klabjan \\
  Industrial Engineering and Management Sciences \\
  Northwestwern University \\
  Evanston, IL 60201 \\
  \texttt{d-klabjan@northwestern.edu} \\
  \And
 Mark Shapiro \\
  Elevance Health, Inc. \\
  Indianapolis, IN 46204 \\
  \texttt{mark.shapiro@carelon.com} \\
}

\begin{document}
\maketitle
\begin{abstract}
Feature acquisition algorithms address the problem of acquiring informative features while balancing the costs of acquisition to improve the learning performances of ML models. Previous approaches have focused on calculating the expected utility values of features to determine the acquisition sequences. Other approaches formulated the problem as a Markov Decision Process (MDP) and applied reinforcement learning based algorithms. In comparison to previous approaches, we focus on 1) formulating the feature acquisition problem as a MDP and applying Monte Carlo Tree Search, 2) calculating the intermediary rewards for each acquisition step based on model improvements and acquisition costs and 3) simultaneously optimizing model improvement and acquisition costs with multi-objective Monte Carlo Tree Search. With Proximal Policy Optimization and Deep Q-Network algorithms as benchmark, we show the effectiveness of our proposed approach with experimental study. 
\end{abstract}


\section{Introduction}
Many machine-learning algorithms work with the assumption that all features have been observed and available during training and testing times or the missing data are disregarded as unacquired. Feature acquisition, a process in which further relevant data are acquired at variable costs, addresses this assumption to more closely align with some real-world applications, Huang \cite{ThirdPaper}. For medical diagnostic tasks, from the basis of incomplete features, doctors sequentially obtain additional test results until they obtain sufficient information to make adequate diagnoses of the patients. Determining which features to acquire is dependent on the previous diagnostic observations and the sequence at which the features are obtained can vary from patient to patient. Although accurate diagnoses are more likely with additional features, acquiring them incurs variable costs and is balanced with the improvement in performance, Melville \cite{originalPaper}. \\
\\
Previous studies on the feature acquisition problem address the trade-off between acquisition costs and performance improvement and the sequential decision making process, and are categorized into non-reinforcement learning and reinforcement learning (RL) approaches. Non-RL approaches focus on selecting the most informative features to acquire based on their utility values. These methods, Melville \cite{originalPaper}, desJardins \cite{CFA}, and Huang \cite{ThirdPaper}, estimate the expected utility of a feature for improving the model performance and acquire the feature with maximum expected utility. Although these methods provide a framework for feature acquisition based on utility values, they focus on subsets of features to acquire at a time, do not consider acquisition costs, or treat the model performance and acquisition costs as an aggregated single objective. RL approaches, Contardo \cite{sequential}, Shim \cite{NextPaper}, and Li \cite{surrogate}, formulate the feature acquisition problem as a Markov decision process (MDP), where the state is the set of currently acquired features and the action is the acquisition of the next feature, and learn the best feature acquisition policy. For each acquisition step, the acquisition cost is incurred and defined as the reward for the action. Prediction error is calculated when the episode ends or the agent decides to stop the acquisition process. Additionally, the additive constraint of the acquisition costs to the rewards also necessitates further fine-tuning of a regularization parameter. \\
\\
Monte Carlo Tree Search, Kocsis \cite{MCTS}, for feature acquisition has the advantage over other RL algorithms in the fact that the reward (prediction) is obtained only at the end of an episode. Our Monte Carlo Tree Search (MCTS) approach also considers intermediary rewards for each acquisition step. We model the reward for each feature acquisition action as the division of the classification prediction probability with the feature being acquired by the cumulative incurred acquisition costs. The cumulative incurred acquisition costs are normalized by the cost of all features. \\
\\
We also propose the trade-off between acquisition costs and model performance as a multi-objective optimization (MO) problem. In MO-MCTS, we model the costs and classification prediction probabilities as two conflicting objectives to be optimized simultaneously. Previous studies have applied the RL algorithms on the additive scalar aggregation of the two objectives. As the objectives are conflicting, the two policies may be incomparable and the Pareto optimal set of solutions need to be found, Wang \cite{MO}. We modify the algorithm presented in Wang \cite{MO} to find the Pareto optimal solution for each feature acquisition step and incorporate it within MCTS. \\
\\
In comparison to the Proximal Policy Optimization, Schulman \cite{PPO}, and Deep Q-Network, Mnih \cite{DQN}, algorithms, our Monte Carlo Tree Search approach shows performance improvements in all the data sets we considered, with the relative improvement in the range of $1.2\%$ to $25.1\%$. The multi-objective Monte Carlo Tree Search implementation shows an advantage in tight budget situations, as it leads to more variable feature acquisition sequences and can thus satisfy different cost budgets and confidence thresholds. \\
\\
Our main contributions in this work are as follows.
\begin{itemize}
  \item We propose to apply Monte Carlo Tree Search (MCTS) for the first time to the feature acquisition problem.
  \item We apply multi-objective MCTS to optimize feature acquisition costs and classification prediction probabilities simultaneously.
  \item We show the advantages of our proposed approaches on three medical data sets and the MNIST data set in comparison to two reinforcement learning approaches (Proximal Policy Optimization in Schulman \cite{PPO} and Deep Q-Network in Mnih \cite{DQN}).
\end{itemize}
Related works are reviewed in Section \ref{related}. Section \ref{FA} presents our approaches in detail. Experimental setup and results are presented in Section \ref{Exp}.
\section{Related Works}\label{related}
\paragraph{Feature Acquisition}
With data sets consisting of incomplete data, ML models typically are trained and utilized with the incomplete data ignored. Selecting the most informative features to acquire, while balancing the costs of acquisition, is important to select real-life situations and to increase the performance of the models. Previous non-RL approaches address the feature acquisition problem from the expected utility of an unacquired feature. Melville \cite{originalPaper} quantifies an Uncertainty Score for a feature, which is defined as the absolute difference between the estimated class probabilities of the two most likely classes when trained with the feature. desJardins \cite{CFA} calculates a Confidence Score for a subset of features based on an ensemble of classifiers. Huang \cite{ThirdPaper} incorporates an iterative supervised matrix completion algorithm with the variance of a feature after the iterations as its utility. Melville \cite{originalPaper} does not consider acquisition costs, but others incorporate them by first sorting the unacquired features by costs, desJardins \cite{CFA}, or constructing an objective function with the acquisition costs and applying gradient descent, Huang \cite{ThirdPaper}. The algorithm stops when all missing features have been acquired or the classification confidence reaches a pre-defined threshold value. RL approaches formulate the problem as a MDP. Contardo \cite{sequential} applies PPO, Schulman \cite{PPO}, to the policy network. Similarly, Shim \cite{NextPaper} also considers the Deep-Q Network, Mnih \cite{DQN}, to model the feature acquisition policy. Li \cite{surrogate} uses a pretrained surrogate model to estimate both the state transitions and the prediction in a unified model in which the intermediate prediction errors based on information gain are also calculated. These RL algorithms provide the sequential feature acquisition steps in which one informative feature is acquired at a time, and the cost and performance trade-off is incorporated into the cumulative rewards. However, the classification errors and acquisition costs are additively aggregated into a single objective function. With the exception of Li \cite{surrogate}, prediction errors are also only calculated at the end of an episode.
\paragraph{Monte Carlo Tree Search}
Monte Carlo Tree Search (MCTS) is first proposed as an algorithm to find near-optimal solutions for large state-space MDPs, Kocsis \cite{MCTS}. By applying the Upper Confidence Bounds (UCB) bandit algorithm, Auer \cite{auer}, MCTS iteratively searches the state space while balancing the exploration of suboptimal actions and exploitation of optimal actions, Kocsis \cite{MCTS}. Given its application to problems that can be formulated as MDPs, it has since been applied to various domains in artifical intelligence, \'Swiechowski \cite{MCTSreview}. AlphaGo and its variants also utilize a neural network in conjunction with MCTS. This network outputs a vector of move probabilities and a scalar value estimation from the position state $s$ and is used as both policy and value networks. The network is then used to guide the Monte Carlo simulations and is iteratively trained using the results from self-play, Silver \cite{AlphaGo}. In our approach, we consider the default uniform random policy for the Monte Carlo simulations and similarly consider iteratively training the acquisition policy based on the simulations. 
\paragraph{Multi-objective Monte Carlo Tree Search}
For multi-objective reinforcement learning problems, previous approaches have focused on optimization based on the total order of the solutions and aggregation of the vectorial objectives into a scalar objective function. Similar to the previous RL approaches, weighted summation of the different objectives has been a popular choice, Wang \cite{MO}. For conflicting objectives, this strategy does not lead to an optimal policy, as there exists a set of optimal solutions ordered along the Pareto Front, Wang \cite{MO}. For the Monte Carlo Tree Search algorithm applied to multi-objective optimization, proposed algorithms have focused on scalarization schemes for the vectorial rewards so that the solutions reach the Pareto Front and the UCB algorithm can be applied. Wang \cite{MO} proposes a hypervolume indicator based scalarization scheme, where the rewards maximizing the indicator belong to the Pareto Front, Fleischer \cite{hv}. Painter \cite{convex} provides a linear transformation scheme to achieve scalarization. In our approach, we closely follow the algorithm in Wang \cite{MO}.
\section{Feature Acquisition using Monte Carlo Tree Search}\label{FA}
\subsection{Problem Statement}
Consider a predictive task with feature vector $X \in \mathbb{R}^d$ and class $y$. For $C \in \{1,\cdots,d\}$, we denote vector $X_C = (X_i)_{i \in C}$. Starting from an empty set of features, we perform a sequential feature acquisition process. We address the case where we obtain complete information with all the features acquired for their ground-truth values. The aim of the process is to obtain the sequences of feature acquisition steps that maximize the task performance while minimizing the acquisition costs. \\
\\
We formulate the problem as a Markov decision process
\begin{align*}
s_t &= X_{O_t},\\
a_t &\in A_t = \{1,\cdots,d\} \setminus O_t, \\
r_t &= \frac{P(\hat{y}|X_{O_t \cup \{a_t\}})}{\frac{\sum\limits_{i=0}\limits^{t}C_{i}}{C_{\text{total}}}}.
\end{align*}
We consider episodic solutions from the empty set of features ($t=0$) to the complete set of features ($t=d$). At a given time, the agent is in state $s_t$ and selects a feature to acquire ($a_t$) according to its policy. The agent then receives the reward $r_t$ from the environment and transitions to the state $s_{t+1} = X_{O_t \cup \{a_t\}}$. The goal of the agent is to maximize the cumulative rewards. 
\paragraph{State} The state at time $t$, $s_t$, is the $X_{O_t}$, the values of the already acquired feature subset $O_t \subseteq \{1, \cdots, d\}$. 
\paragraph{Action} The action space at time $t$ is the unacquired feature set $A_t$. The action at time $t$ is then the acquisition step for a candidate feature with its value $X_{a_t}$. 
\paragraph{Reward} The reward at all times of the episode is defined as the fraction of the classification prediction probability and the normalized incurred acquisition costs up to time $t$. The prediction is made with the feature vector consisting of the acquired feature subset $X_{O_t \cup \{a_t\}}$. The incurred acquisition costs $\smash[b]{\sum\limits_{i} C_{i}}$ is normalized by the total cost $C_{\text{total}}$ of all features. 
\subsection{Monte Carlo Tree Search for Feature Acquisition}
We present the Upper Confidence Tree MCTS algorithm, Kocsis \cite{MCTS}, with our approach-specific implementation details. Starting from an empty feature state as the root node, MCTS explores and builds a search tree with $N$ simulations. Each simulation consists of three phases, \'Swiechowski \cite{MCTSreview}. 
\paragraph{Selection:} Starting from the root node, a feature is selected iteratively until arriving at a leaf node. The set $A_{s_t}$ of admissible features in node/state $s_t$ defines the child nodes of $s_t$. Feature selection according to the maximization of the Upper Confidence Bound, Auer \cite{auer}, reads
\begin{align}
a^{*}_t = \argmax\limits_{a_t \in A_{s_t}} \, Q(s_t,a_t) + c\sqrt{\text{ln}(n_{s_t})/n_{s_t,a_t}},
\end{align}
where $Q(s_t,a_t)$ is the average cumulative reward of feature $a_t$, $n_{s_t}$ is the visit count of node $s_t$, and $n_{s_t,a_t}$ is the number of times $a_t$ has been selected in node $s_t$. The exploration and exploitation trade-off is controlled by the hyperparameter $c$, which is optimized as described in a next section. 
\paragraph{Expansion:} Once a leaf node has been selected, all the absent child nodes of the leaf node are added to the tree. 
\paragraph{Simulation:} Starting from the leaf node, a feature is selected uniformly at random until the terminal state is reached. Differently from previous studies in AlphaGo and its variants, we utilize the uniform random policy as our default simulation policy. As defined in the previous section, we compute the reward for each feature and calculate the cumulative reward.
\paragraph{Backpropagation:} During backpropagation, $Q(s_t,a_t)$, $n_{s_t,a_t}$, and $n_{s_t}$ are updated
\begin{align*}
r_{s_t,a_t} &= \sum\limits_{t'=t}\limits^{d} r_{t'}, \\
n_{s_t,a_t} &= n_{s_t,a_t} + 1, \\
n_{s_t} &= n_{s_t} + 1, \\
Q(s_t,a_t) &= \frac{r_{s_t,a_t}}{n_{s_t,a_t}}. 
\end{align*}
After $N$ simulations and updated statistics using backpropagation, the feature acquisition action is defined as
\begin{align} \label{eq:1}
a^{*}_t = \argmax\limits_{a_t \in A_{s_t}} \, Q(s_t,a_t).
\end{align}
The next state is then obtained according to the acquisition step and $N$ further simulations are conducted with the next state as the new root node. This process continues until the terminal, complete feature state is reached. \\
\\
We have two variants of the MCTS algorithm. In the \textbf{standalone} implementation, we conduct MCTS training by constructing a search tree for each sample in the training data set. The visited states and their $Q$ values are then stored for the entire training data set. This stored set is then used to calculate the next feature probabilities for each visited state. The next feature probabilities are calculated with the cumulative $Q$ values for each admissible feature. We then train a policy network with the visited states and their next feature probabilities. \\
\\
In the \textbf{integrated} implementation, we embed a policy network in the training phase and periodically train the network during MCTS training. After initializing with random weights, the network is then used to guide the feature acquisition step. The network is periodically trained with visited states and their next feature probabilities. We also optimize the network train frequency. \\
\\
The pseudocodes for our \textbf{integrated} implementation is shown in Algorithm \ref{alg:so_int_pseudocode}. We highlight the problem specific details in embedding the policy network and its training on the visited states and their next feature probabilities. The remaining functions are provided in the Appendix \ref{appendix:so}.
\begin{algorithm}
    \caption{Single-objective Monte Carlo Tree Search (Integrated)}\label{alg:so_int_pseudocode}

    \SetKwInOut{Input}{Input}
    \Input{Iteration number $I$, initial policy network weights $\theta$, policy network update frequency $f$}
    \nonl \;
    Initialize policy network $\phi$ with $\theta$\\
    Initialize list $L$ of visited nodes and their $Q$ and visit counts $N$\\
    $i \leftarrow 0$ \\
    \nonl \;
    \begin{multicols}{2}
    \textbf{for} sample = 1,2,$\ldots$,$m$ \textbf{do} \\
    \qquad $i \leftarrow i+1$ \\
    \qquad Initialize state $s_0$ \\
    \qquad Create root node $v_0$ with $s_0$ \\
     \qquad \qquad $Q(v_0)$: reward of $v_0$ \\
     \qquad \qquad $N(v_0)$: visit count of $v_0$ \\
     \qquad \qquad $C(v_0)$: children of $v_0$ \\
     \qquad \qquad $a(v_0)$: action of $v_0$ \\
     \qquad \textbf{while} $v_0$ not terminal \textbf{do}\\
     \qquad \qquad \textbf{MCTS}$(v_0,$I$)$\\
     \qquad \qquad $a$ $\leftarrow$ $\phi_{\theta}$($s_0$)\\
     \qquad \qquad $v_0 \leftarrow \textbf{makeChild}(v_0,a)$\\
     \qquad \textbf{end while} \\
     \qquad Append $Q(v)$ and $N(v)$ for $v$ in \textbf{MCTS} to $L$\\
     \qquad \textbf{if} $f$ $\%$ $i$ == 0 \textbf{do}\\
     \qquad \qquad $S$, $A$ $\leftarrow$ \textbf{preprocess}($L$)\\
     \qquad \qquad Train $\phi_{\theta}$ on $S$ and $A$\\
     \qquad \textbf{end if} \\
     \textbf{end for}\\
    \vfill\null
    \columnbreak
    \nonl\;
    \nonl\;
    \nonl\;
    \nonl\;
    \nonl\;
     \underline{function} \textbf{preprocess}($L$)\\
     \qquad Make each node $v$ in $L$ to be distinct with addition for $Q(v)$ and $N(v)$ for duplicates\\
     \qquad $A$ = $\vv{0}$\\
     \qquad $S$ = $v$ in $L$ \\
     \qquad \textbf{for} $v$ in $L$ \textbf{do} \\
    \qquad \qquad \textbf{for} action in $A$ \textbf{do} \\
    \qquad \qquad \qquad Find child nodes of $v$ in $L$ \\
    \qquad \qquad \qquad \textbf{for} node in child nodes \textbf{do} \\
    \qquad \qquad \qquad \qquad $A(\text{action})$ $\mathrel{+}=$  $Q(\text{node})$/$N(\text{node})$\\
    \qquad \qquad \qquad \textbf{end for}\\
    \qquad \qquad \textbf{end for}\\
    \qquad Normalize $A$ with division by max($A$)\\
    \qquad \textbf{return} $S$, $A$\\
\end{multicols}
\end{algorithm}
\subsection{Feature Acquisition using Multi-objective Monte Carlo Tree Search}
In this section, we present the multi-objective-MCTS algorithm in Wang \cite{MO} with our modifications in the reward formulation and scalarization, and Pareto Front approximation. 
\begin{algorithm}
    \caption{Multi-objective Monte Carlo Tree Search}\label{alg:mo_pseudocode}
    \begin{multicols}{2}
    \underline{function} \textbf{expand}($v$) \\
    \qquad \textbf{for} all unacquired actions $a \in$ $A(v)$ \textbf{do} \\
    \qquad \qquad $v'$ $\leftarrow$ $\textbf{makeChild}(v,a)$ \\
    \qquad \qquad Add $v'$ to $C(v)$ \\
    \qquad \qquad $a(v') \leftarrow a$ \\
    \qquad \qquad $R(v')[0] \leftarrow \textbf{classificationProbability}(v')$ \\
    \qquad \qquad $R(v')[1] \leftarrow \textbf{findCost}(v')$ \\
    \qquad \qquad $P$ $\leftarrow$ \textbf{findGlobalP}($P$,$R(v')$)\\
    \qquad \textbf{end for}\;
    \vfill\null
    \columnbreak
    \underline{function} \textbf{findGlobalP}($P$,$R(v)$) \\
    \qquad $P$ $\leftarrow$ $R(v)$ $\cup$ $P$\\
    \qquad Find the non-dominated set $P_{\text{non}}$ in $P$\\
    \qquad $P$ $\leftarrow$ $P_{\text{non}}$\\
    \qquad \textbf{return} $P$\;
    \end{multicols}
    \underline{function} \textbf{simulate}($v$) \\
    \qquad reward = [] \\
    \qquad \textbf{while} $v$ not terminal \textbf{do} \\
    \qquad \qquad Choose $a$ $\in$ $A(v)$ uniform randomly \\
    \qquad \qquad $v$ $\leftarrow$ $\textbf{make child}(v,a)$ \\
    \qquad \qquad R(v)$[0]$ $\leftarrow R(v)$[0]$ + \textbf{classificationProbability}(v)$\\ 
    \qquad \qquad R(v)$[1]$ $\leftarrow R(v)$[1]$ + \textbf{findCost}(v)$ \\
    \qquad \qquad $P$ $\leftarrow$ \textbf{findGlobalP}($P$,$R(v)$)\\
    \qquad \qquad reward$[0]$ $\leftarrow$ reward$[0]$ $ + $ \textbf{classificationProbability}(v)\\
    \qquad \qquad reward$[1]$ $\leftarrow$ reward$[1]$ $ + $ \textbf{findCost}(v)\\
    \qquad \textbf{end while}\;
    \qquad \textbf{return} reward\;
\end{algorithm}
\paragraph{Vectorial Rewards} We define the reward for all timesteps in an episode as the vector of negative normalized incurred acquisition costs and classification probability. During backpropagation, the rewards are updated component-wise as
\begin{align*}
  r_c &= \sum_{t'=t}^{d} r_{t',c}, \\
  r_p &= \sum_{t'=t}^{d} r_{t',p}, 
\end{align*}
where $r_{t',c}$ and $r_{t',p}$ are the negative normalized incurred costs and classification probabilities, respectively.
\paragraph{Pareto Front Approximation} In Wang \cite{MO}, an approximation to the Pareto Front is maintained during training, which we use in the UCB feature selection and feature acquisition policy. When new nodes are added during the expansion and simulation phases, the Pareto Front approximation is updated with the vectors of normalized incurred costs and classification probabilities of the added nodes. We then determine the non-dominated set and denote it as \textbf{P}. We use \textbf{P} as the estimated Pareto Front for the data set. The pseudocode with the modifed expansion and simulation is shown in Algorithm \ref{alg:mo_pseudocode}. The remaining functions are in the Appendix \ref{appendix:mo}.
\paragraph{Reward Scalarization} As in Wang \cite{MO}, we calculate the hypervolume indicator as the reward scalarization method
\begin{align*}
  HV(r;z) = \mu\left(r;z\right),
\end{align*}
which is defined as the Lebesgue measure with respect to a reference point $z$, Fleischer \cite{hv}. Vector $z$ is set at $(-1.0,0)$ so that it is dominated by every $r \in \textbf{P} \cup \{r\}$. Then, the modified Upper Confidence Bounds selection is
\begin{align*}
  Q(s_t,a_t) &= \frac{HV(\textbf{P} \cup \{r\};z)}{n_{s_t,a_t}}, \\
  a^{*}_t &= \argmax\limits_{a_t \in A_{s_t}} \, Q(s_t,a_t) + c\sqrt{\text{ln}(n_{s_t})/n_{s_t,a_t}}.
\end{align*}
For the acquisition policy, the next state is obtained with the selected acquisition feature and serves as the next root node. We also embed the policy network in the training phase in the \textbf{integrated} implementation. 
\section{Experiments}\label{Exp}
\subsection{Data Sets and Benchmark Algorithms}
We use four data sets. 
\textbf{(a) Heart Failure (HF)} from Chicco \cite{Chicco}: This data set contains medical records of $299$ patients who had heart failure with $13$ clinical features and $2$ classes (boolean for death event). \textbf{(b) Coronary Heart Disease (CHD)} from the Framingham Heart Studies Organization \cite{chd}: The Framingham Heart Disease data set contains medical records of $4$,$238$ patients with $16$ risk factors for coronary heart disease as features and the ten year presence of CHD as the class. \textbf{(c) PhysioNet} from Goldberger \cite{Physionet}: The data set from the PhysioNet/CinC Challenge $2012$ consists of medical records of $4$,$000$ ICU stay patients. The data set has $39$ clinical features with $2$ classes for the death event. \textbf{(d) MNIST} from Deng \cite{MNIST}: Each $4\times4$ block is considered as a feature with $70$,$000$ samples, $49$ features, and $10$ classes. \\
\\
For the three medical datasets, acquisition cost is set at $1$ and $7$ for categorical and continuous features, respectively. These costs are determined by the costs of the medical tests required and comparing them to a previous data set where the relative costs of similar tests were quantified, Cestnik \cite{cost}. For the MNIST data set, we also define blocks of $4\times4$ pixels as features. For each block, the acquisition cost is defined as $16$ with $1$ for each pixel.\\
\\
For the experiments, we create $4$ splits and use $3$ seeds for the total of $3$ experimental runs. For each data set, the $80/20$ split is used for the training and test samples. \\
\\
We use Proximal Policy Optimization (PPO) and Deep-Q Network (DQN) as the baseline algorithms to compare to our approaches. For PPO, we also incorporate two variants of the algorithm: PPO-PG and PPO-AC with the difference in network update frequencies to reflect vanilla policy gradient and actor-critic methods, respectively. The network architectures of the algorithms are provided in the Appendix \ref{appendix:network}.
\subsection{Setup and Evaluation Metrics}
\begin{figure}
\centering
\includegraphics[width=0.4\linewidth]{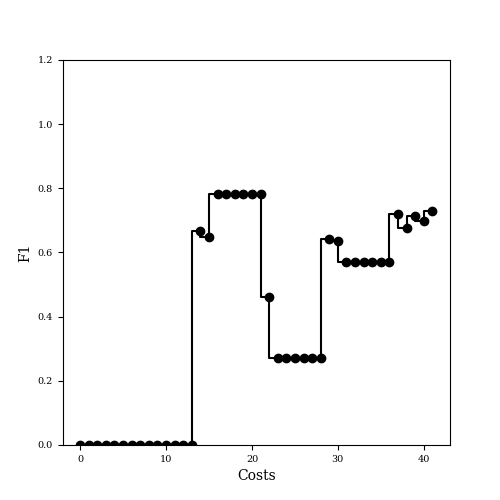}
\caption{F1 score curve on the incurred acquisition costs.}
\label{fig:f1_curve}
\end{figure}

For evaluation of the feature acquisition algorithms, we plot the F1 scores against the incurred acquisition costs. We then calculate the areas under the curves (AUCs) of the resulting F1 curves and average them across the splits and seeds. We also report the highest test F1 AUC values of the $3$ experimental runs of each algorithm. Since the obtained feature acquisition sequences do not contain all the cost points up to the full cost of all features, we also extrapolate the F1 scores at these points with the F1 scores of lower costs that are visited by the solution policy. Figure \ref{fig:f1_curve} shows a sample run of our experiments. All experiments are run on a server with Intel core i$9$-$13900$k and NVIDIA GeForce RTX $3080$ graphics card. \\
\\
We use the logistic regression and neural network classifiers for the calculation of the rewards during training and for the evaluation of the F1 scores. To this end, we utilize the following $4$ classifier strategies. \textbf{Pretrain}: The pretrain strategy uses classifiers trained on complete feature vectors. \textbf{Random}: The classifiers are trained on random subsets of the features. \textbf{Retrain}: Starting with the pretrain strategy, classifiers are retrained on the augmented data set with the feature vectors of states visited during training of the algorithms. The frequency at which the classifiers are retrained is optimized by the resulting AUC of the train F1 curve. \textbf{Fit}: In the fit strategy, each subset of the feature set is used to train a single classifier. Each classifier is used for the same subset of features whose states are visited. This strategy is considered for the HF, CHD, and PhysioNet data sets where the numbers of features are low.\\
\\
For categorical features, the unacquired features are set as its own categories and we one-hot encode such features. For continuous features, we initialize at $-1$ (all feature values in our data sets are non-negative). With the MNIST data set, all the unacquired features are set at $0$; this value is used for the policy networks and classifiers. For other data sets, we also utilize hyperparameters to determine how the values of the unacquired continuous features are set with respect to the acquisition costs in calculating classification prediction probabilities and training the policy networks. Using $0$ at $0$ cost and varying the values at full acquisition cost from $0$ to a large negative value (this hyperparameter is set at $-100$ in our experiments), we fit a quadratic, linear, or constant function with the value at full cost. The best strategy is determined by the resulting AUCs of the train F1 curves for each algorithm and classifier. We then use the identified function for setting the all yet to be acquired continuous features. The optimized functions and values are provided in the Appendix \ref{appendix:hyperparameters}. \\
\\
Hyperparameters in the algorithms were optimized based on the resulting F1 AUCs. For PPO, the number of episodes, entropy and value coefficients and learning rates were optimized. The number of episodes, learning rates and $\epsilon$-decay parameter were optimized in DQN. For MCTS, the number of simulations and UCB parameter were optimized. For the Retrain classifier strategy and the integrated implementations of MCTS, the retrain frequencies were also optimized. The optimized hyperparameters are provided in the Appendix \ref{appendix:hyperparameters}. 
\subsection{Experimental Results}
\subsubsection{F1 AUC}
\begin{figure}
\centering
\includegraphics[width=0.4\linewidth]{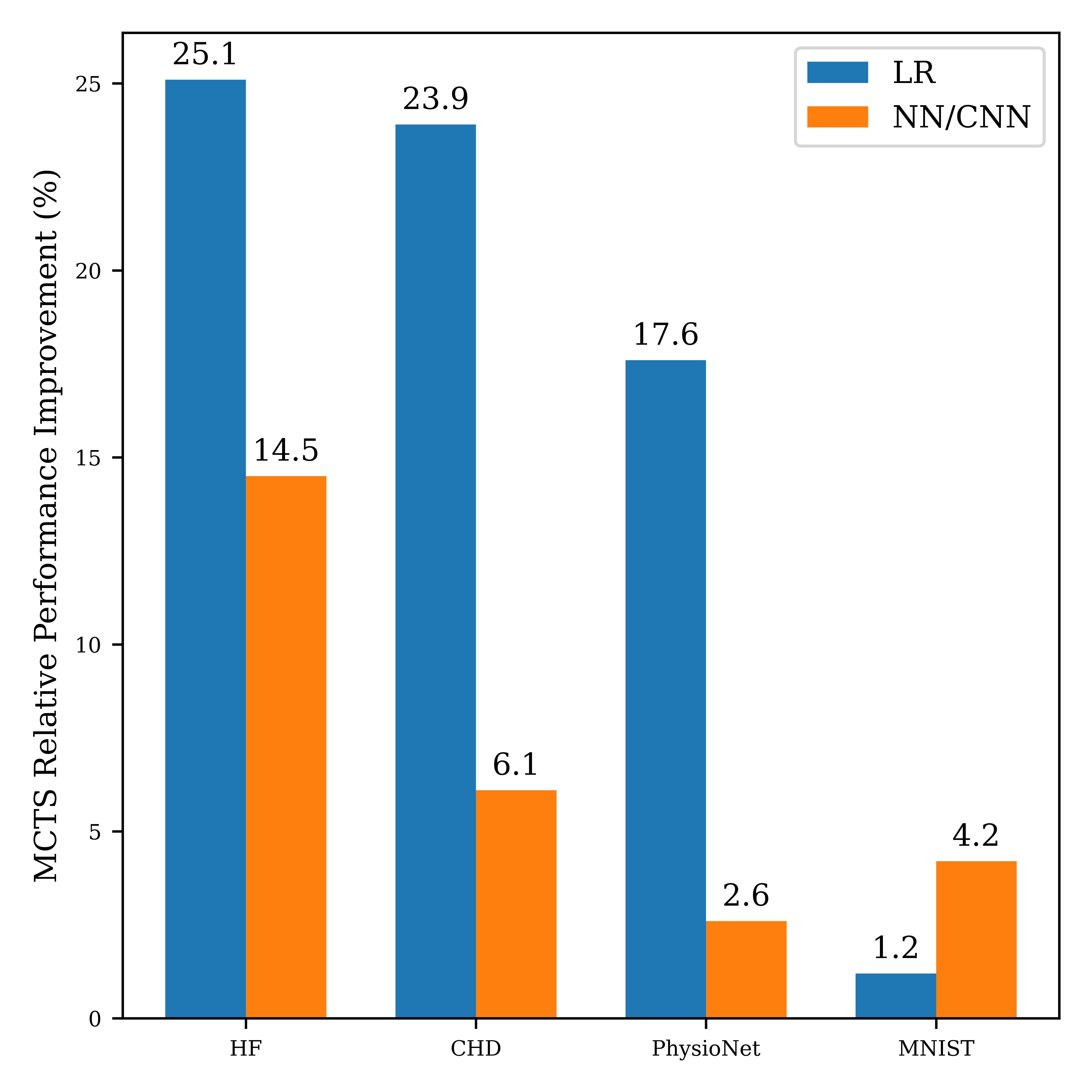}
\caption{Relative differences between the best performing Monte Carlo Tree Search implementation and the benchmark algorithms (LR: logistic regression, NN/CNN: neural network/convolutional neural network).}
\label{fig:bar_chart} 
\end{figure}

\begin{figure}
\centering 
\includegraphics[width=0.19\linewidth]{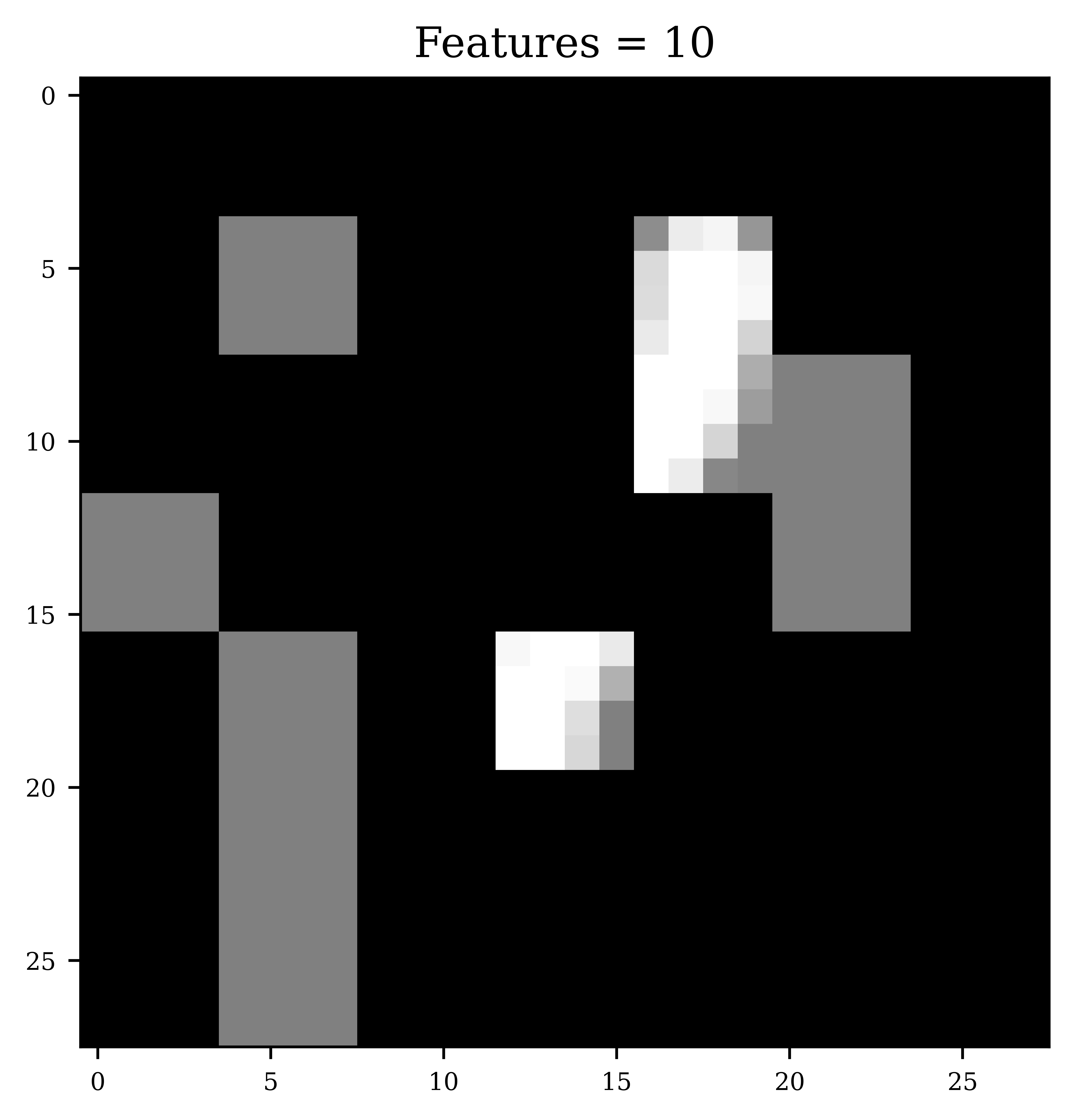}
\includegraphics[width=0.19\linewidth]{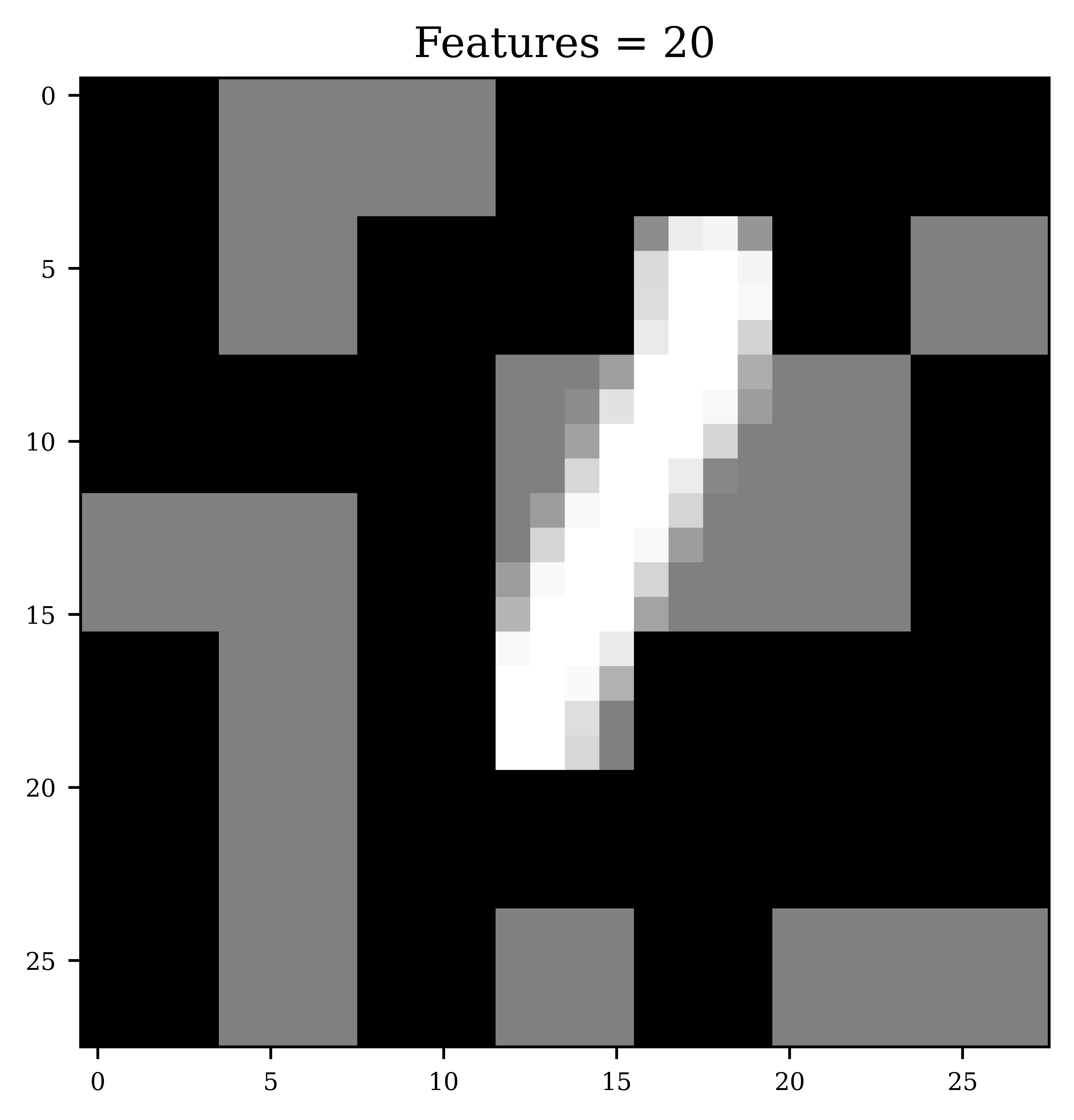}
\includegraphics[width=0.19\linewidth]{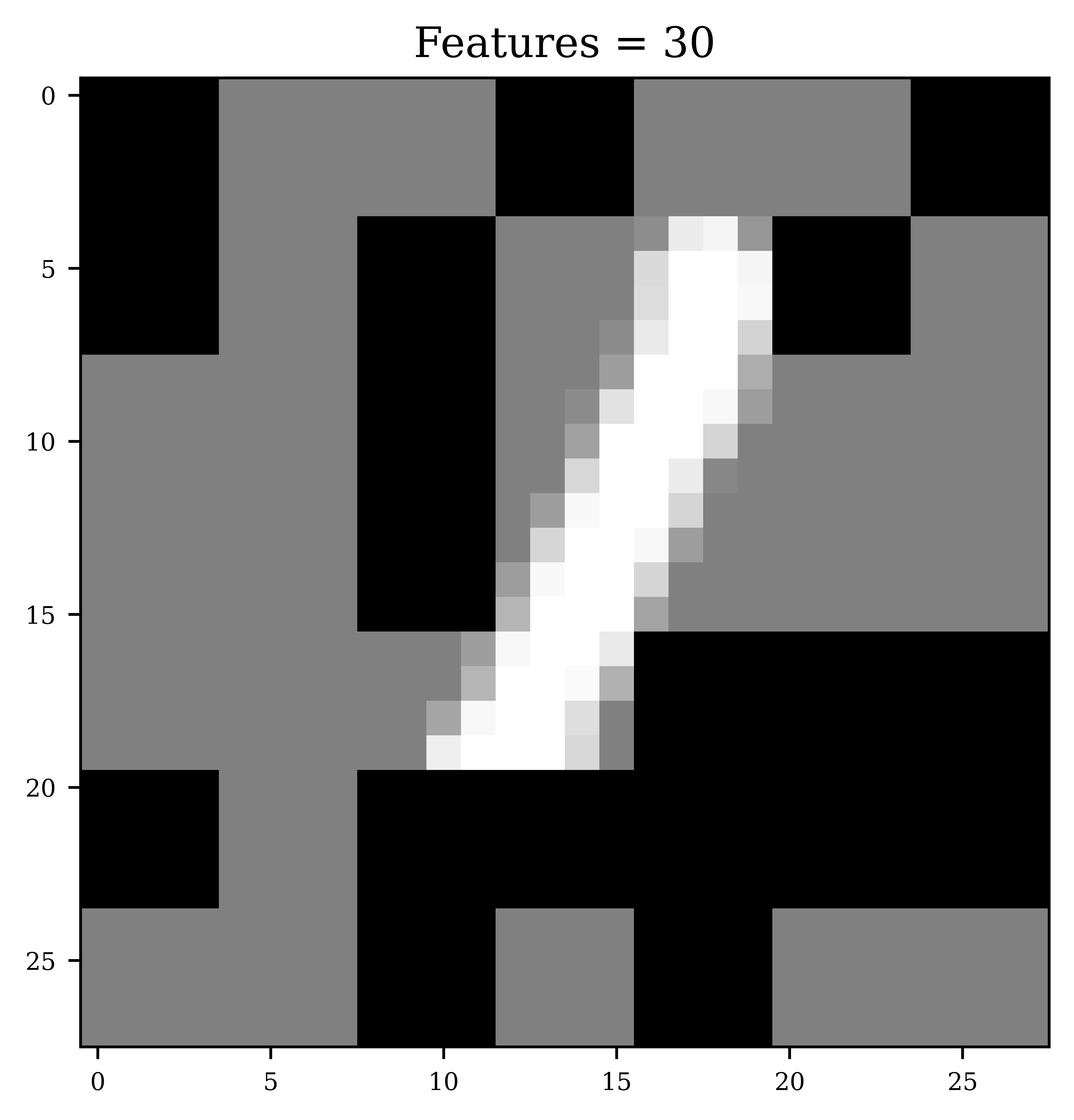}
\includegraphics[width=0.19\linewidth]{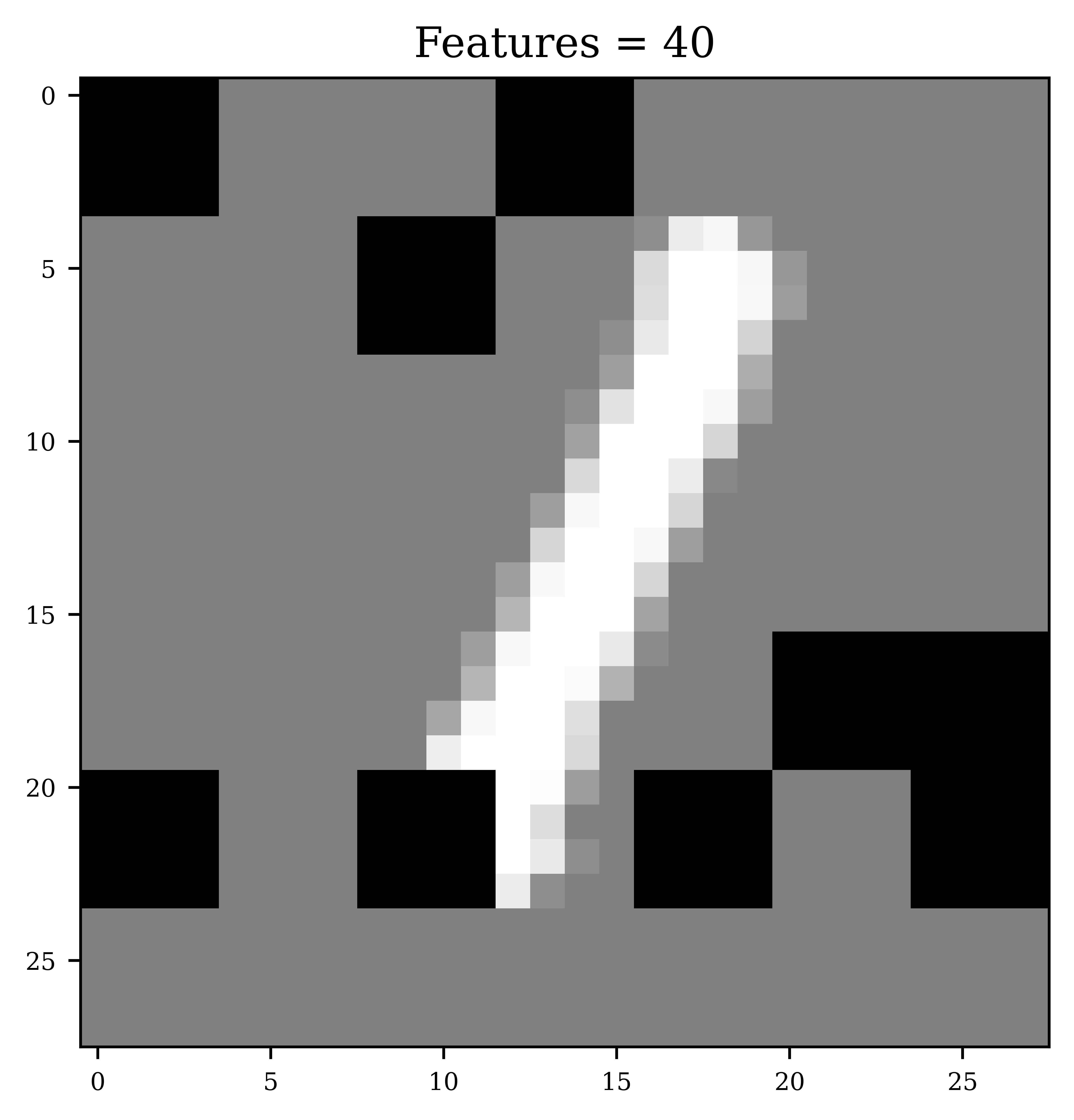}
\includegraphics[width=0.19\linewidth]{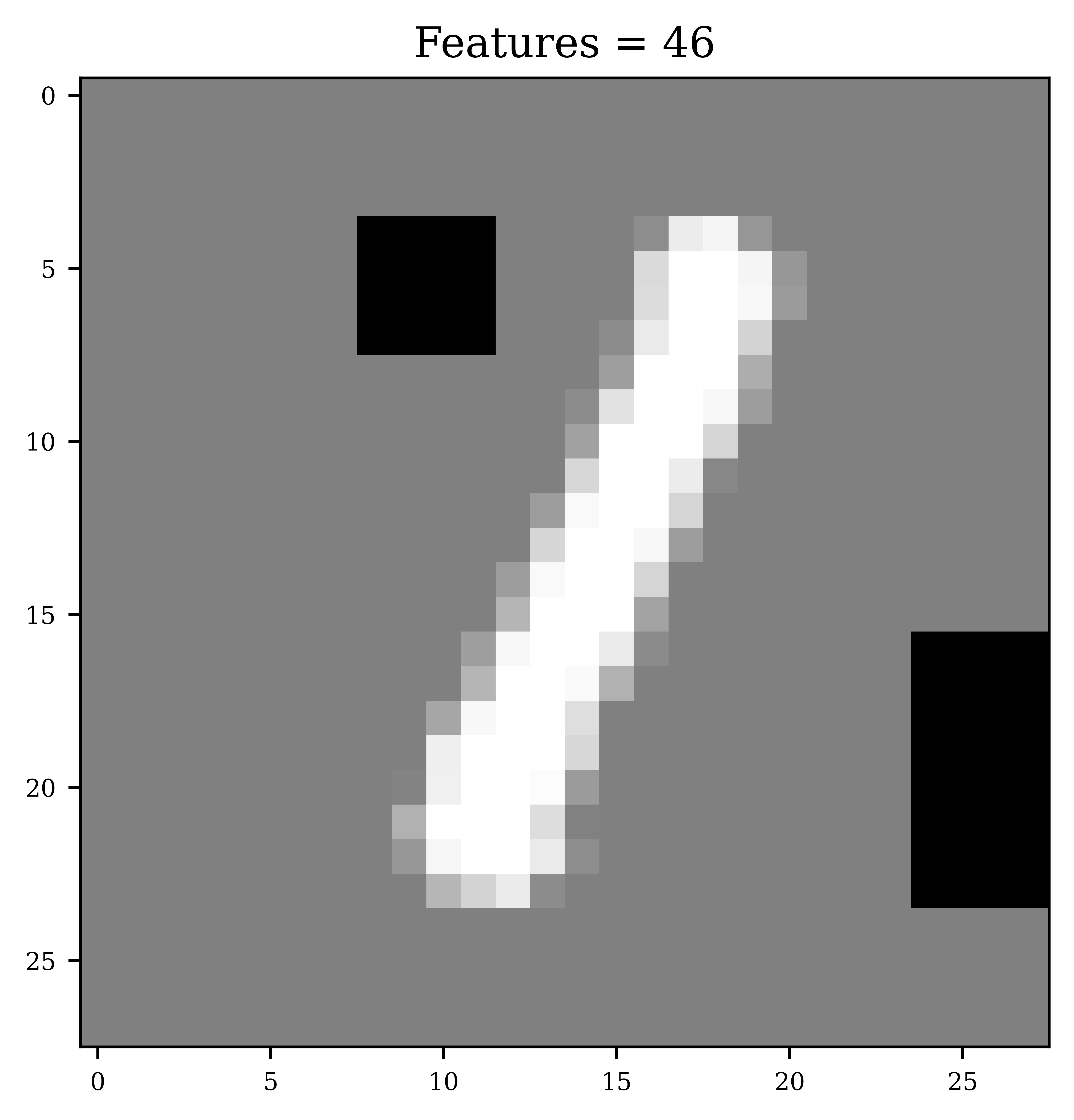}
\includegraphics[width=0.19\linewidth]{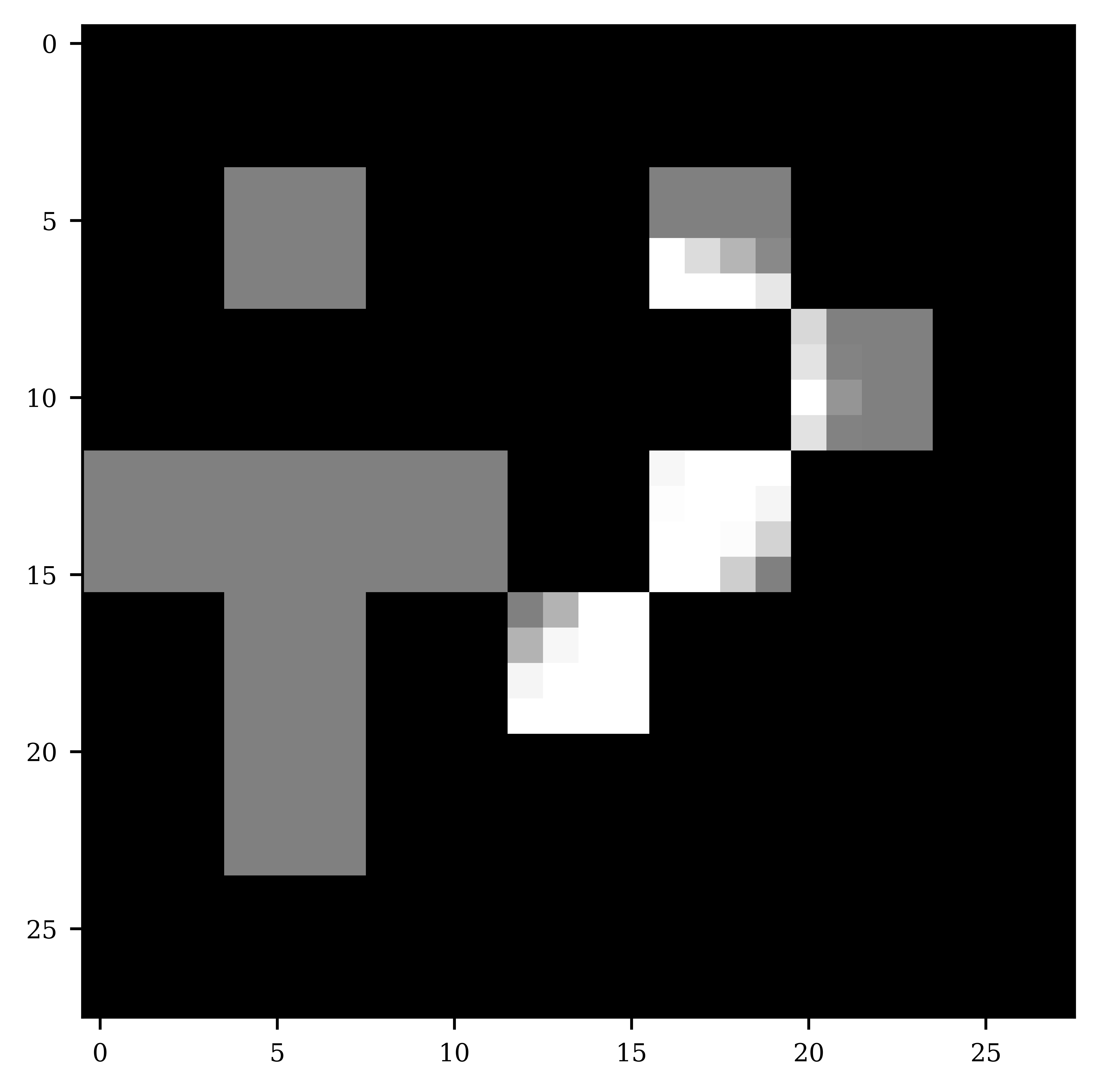}
\includegraphics[width=0.19\linewidth]{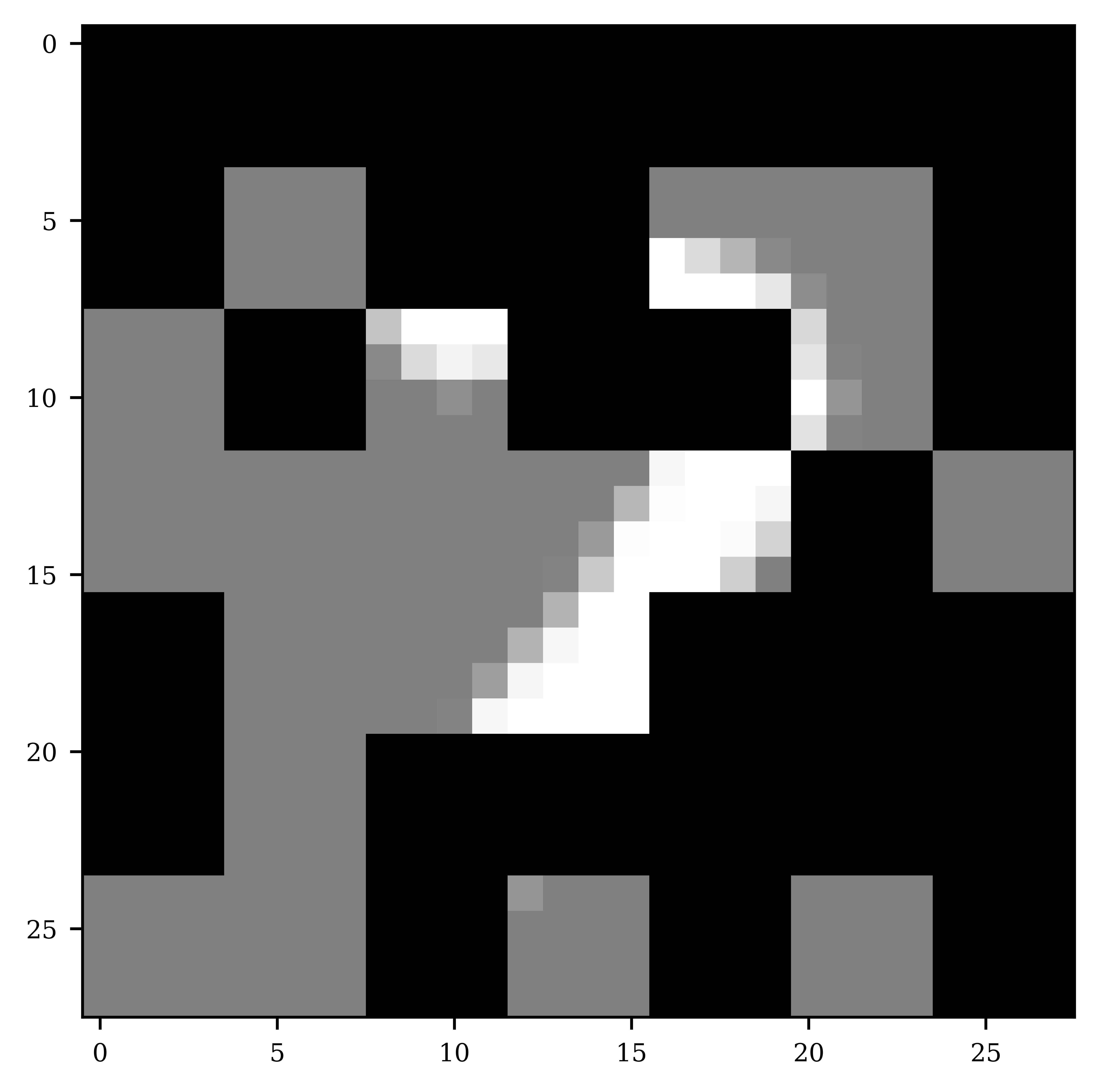}
\includegraphics[width=0.19\linewidth]{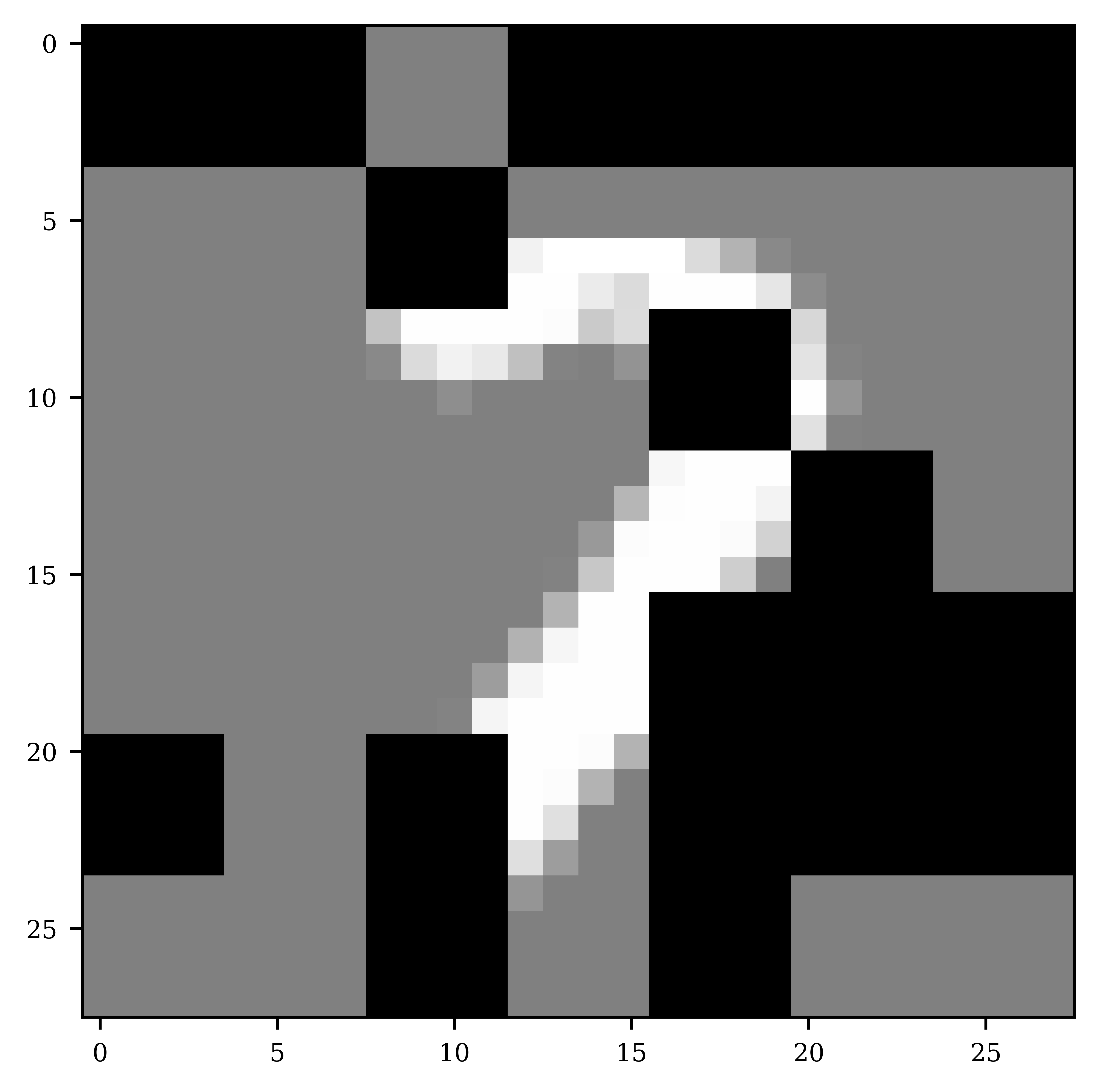}
\includegraphics[width=0.19\linewidth]{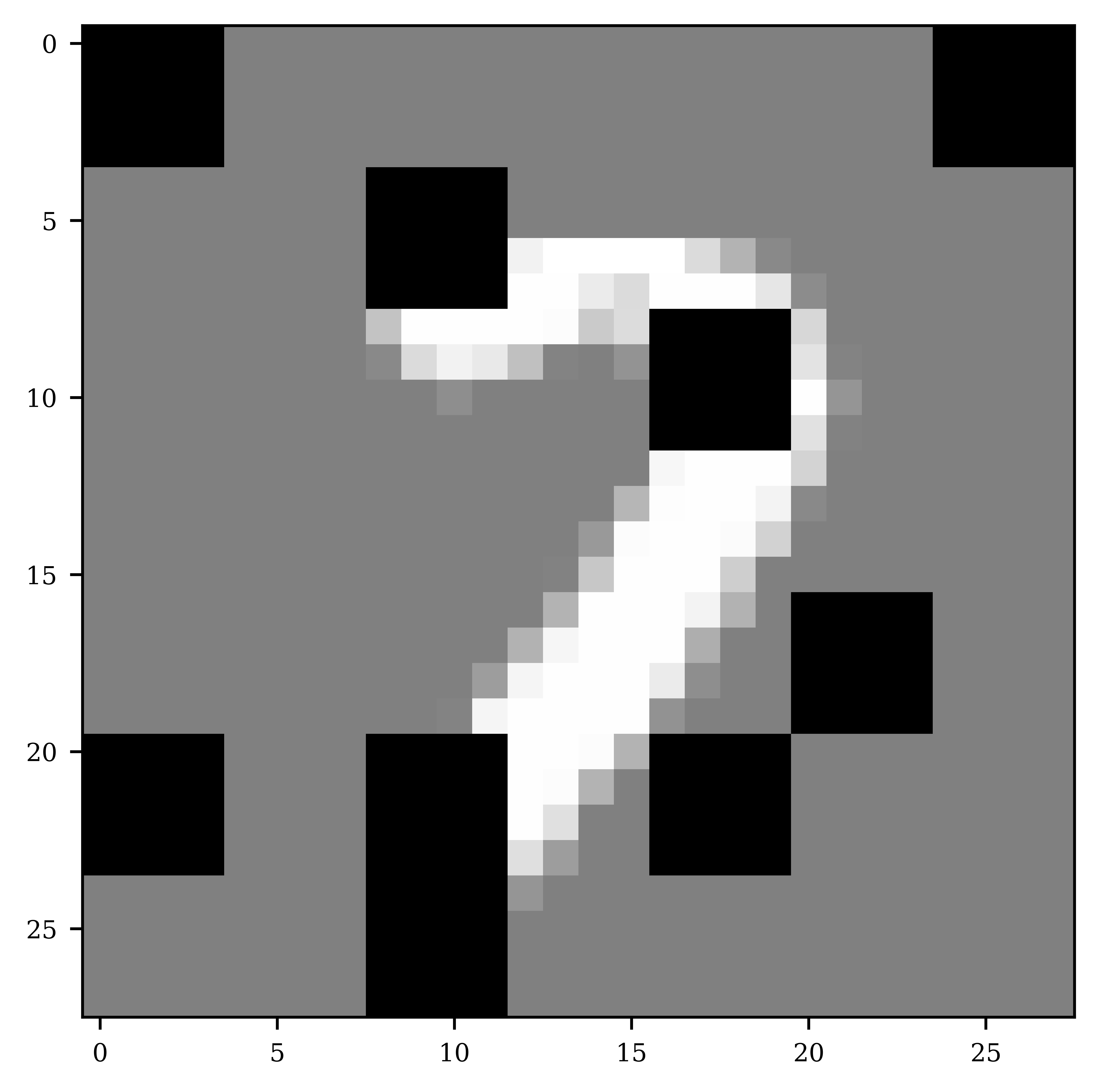}
\includegraphics[width=0.19\linewidth]{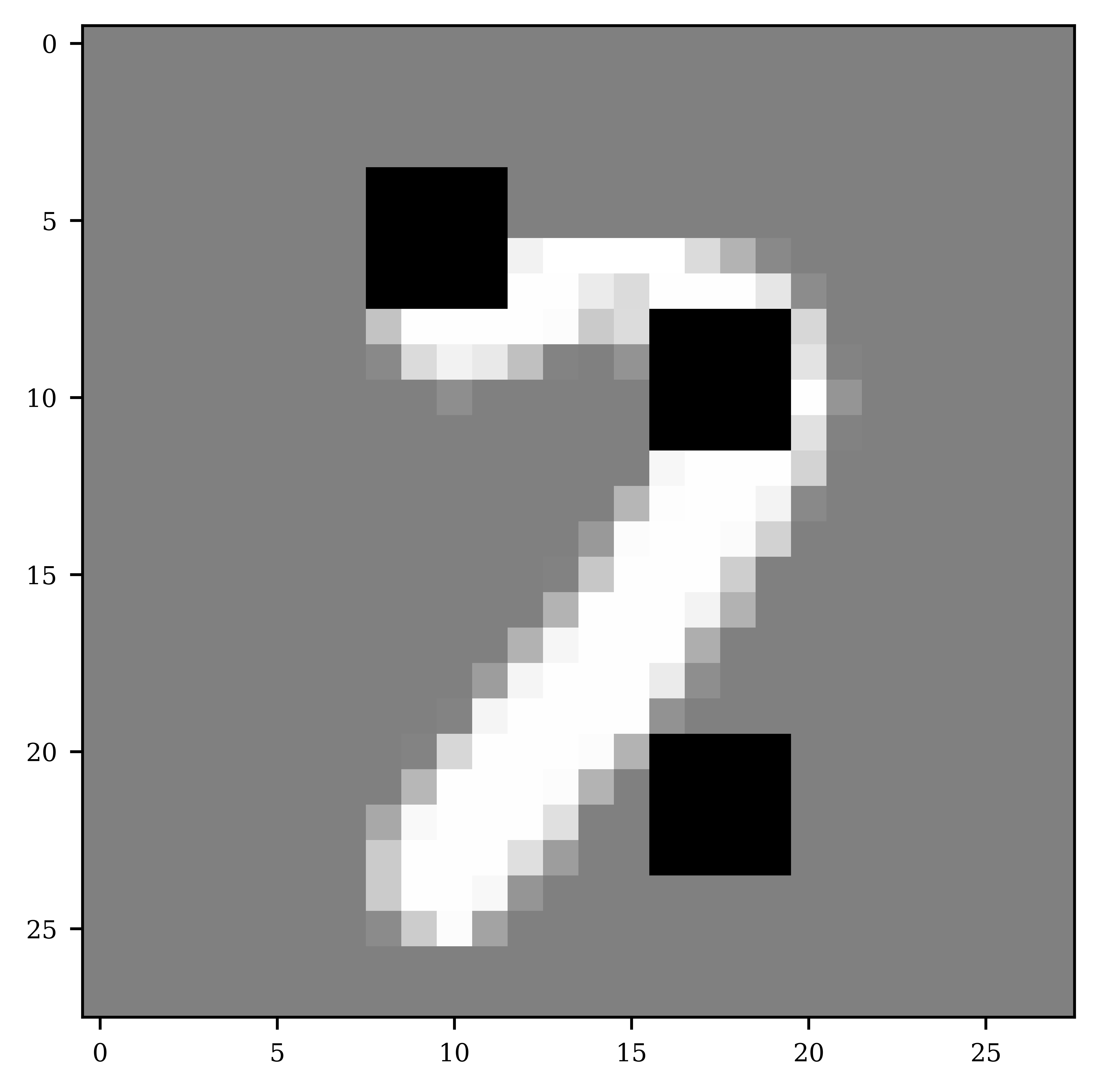}
\includegraphics[width=0.19\linewidth]{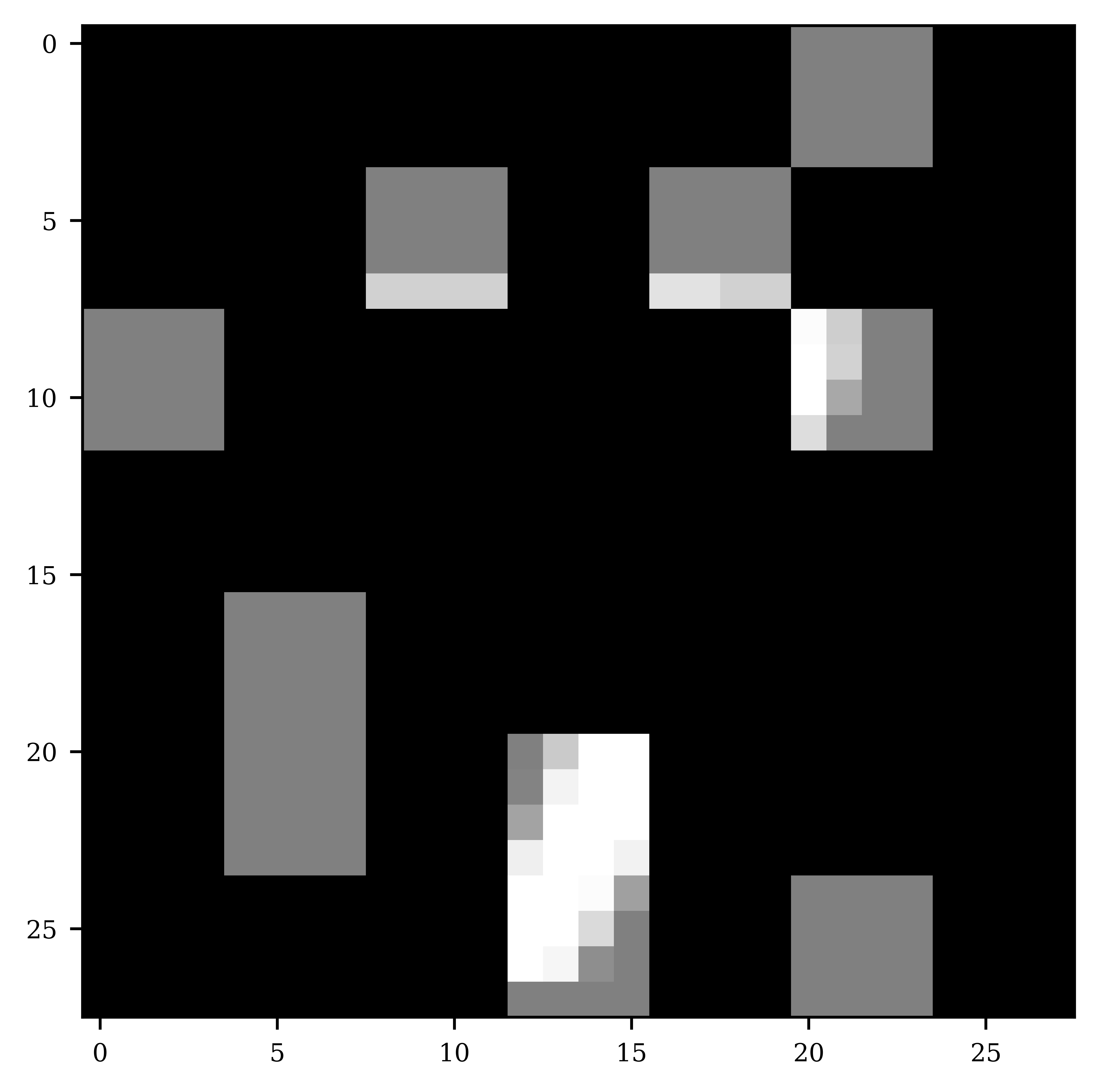}
\includegraphics[width=0.19\linewidth]{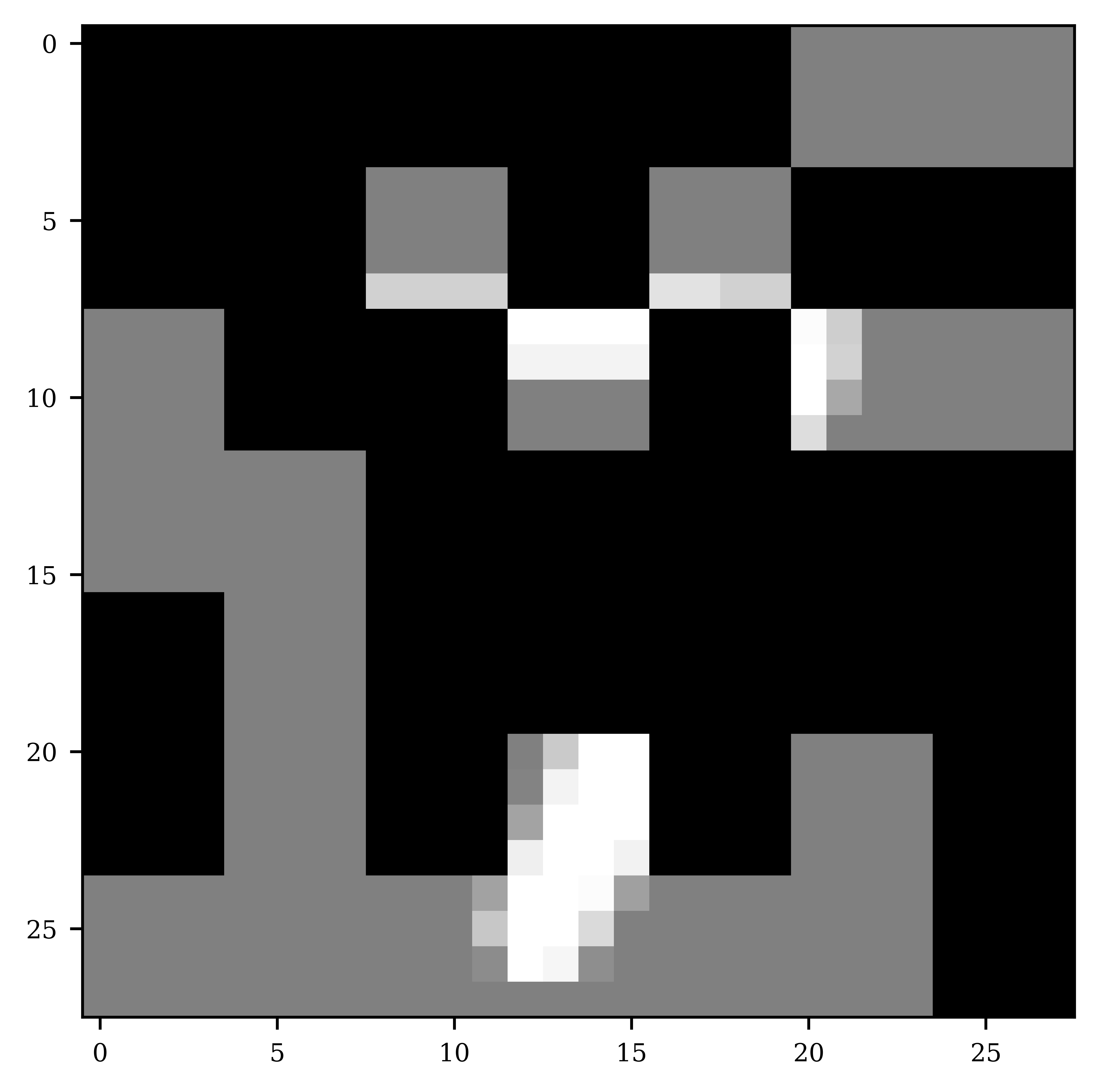}
\includegraphics[width=0.19\linewidth]{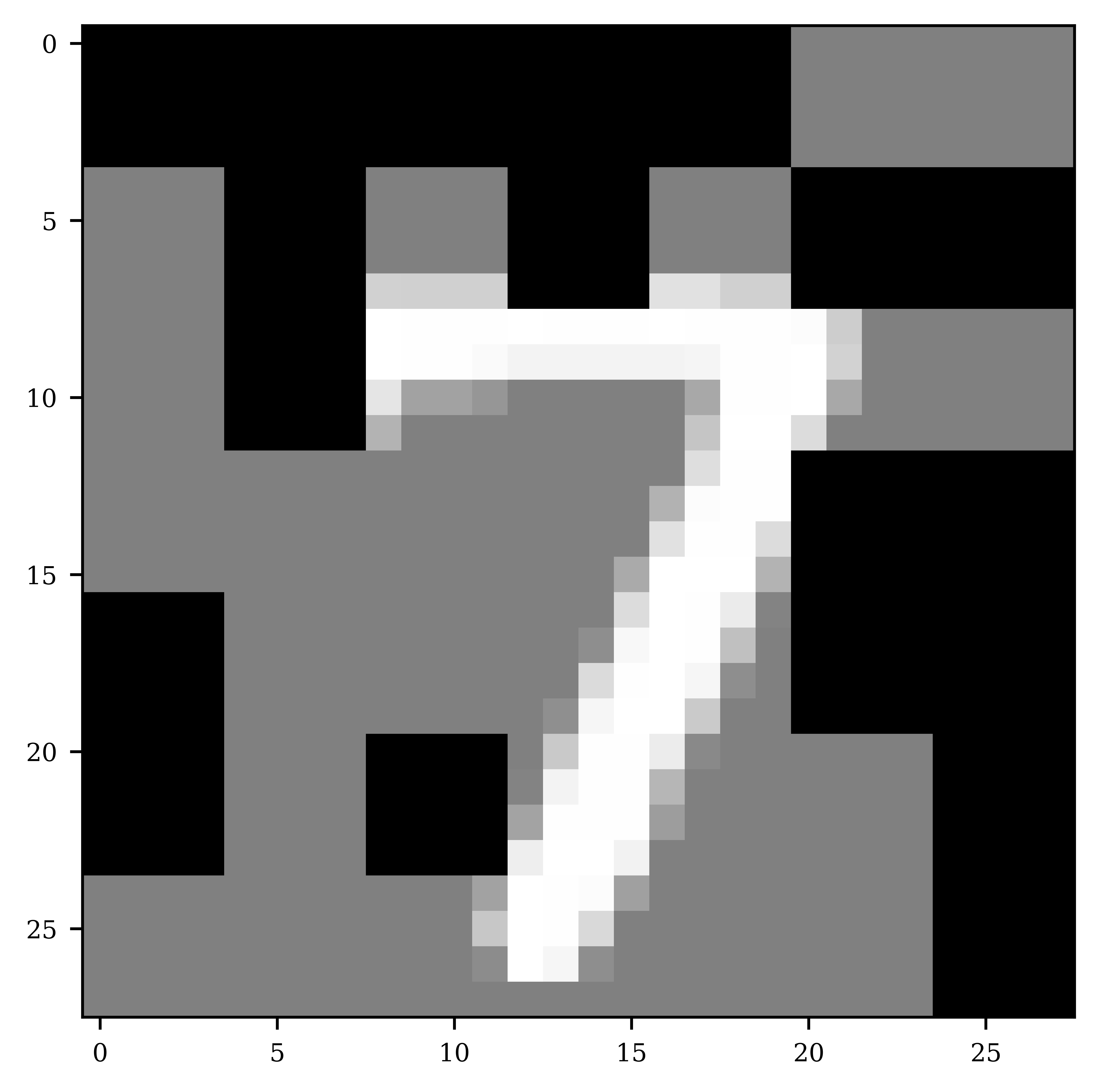}
\includegraphics[width=0.19\linewidth]{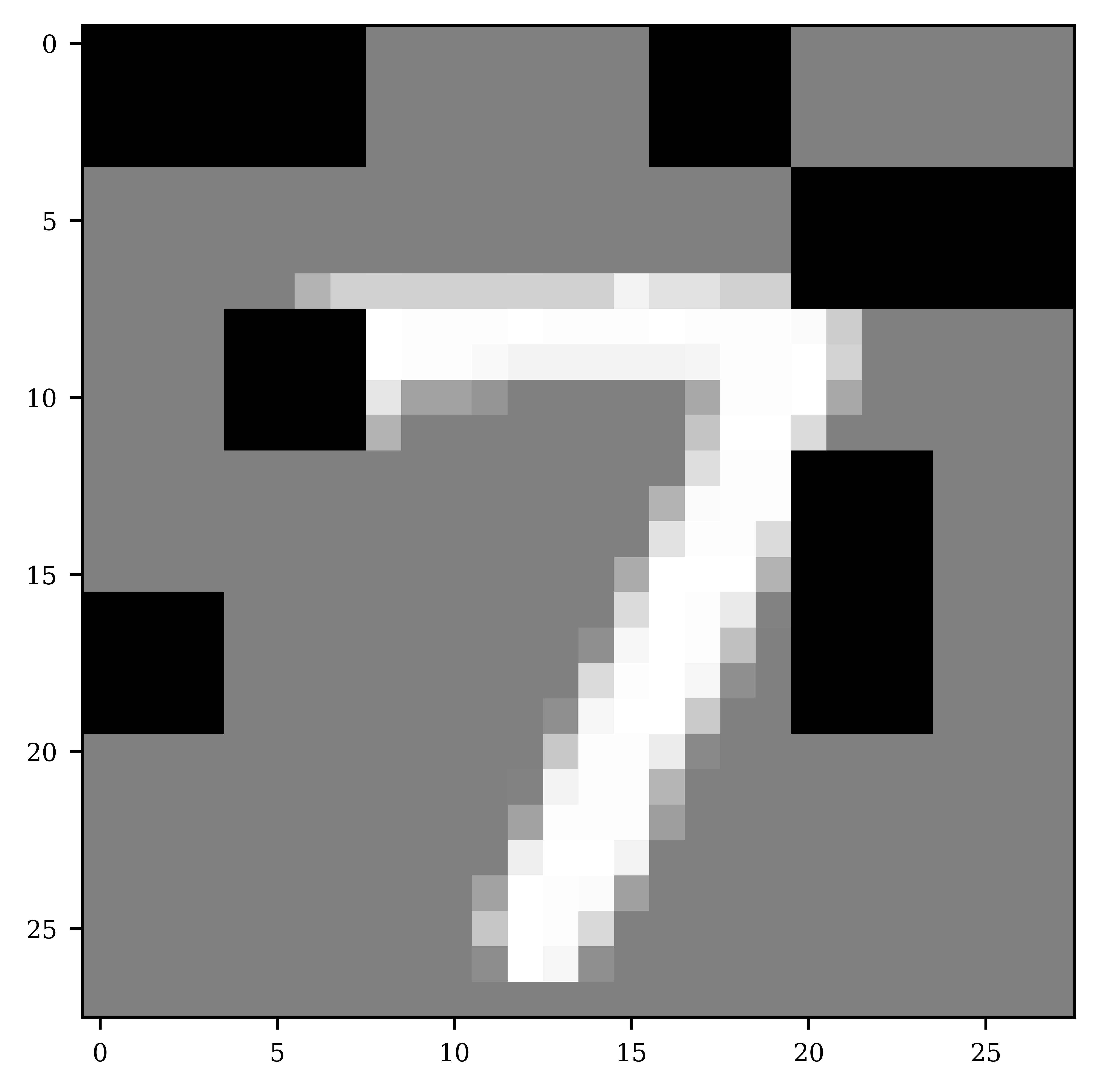}
\includegraphics[width=0.19\linewidth]{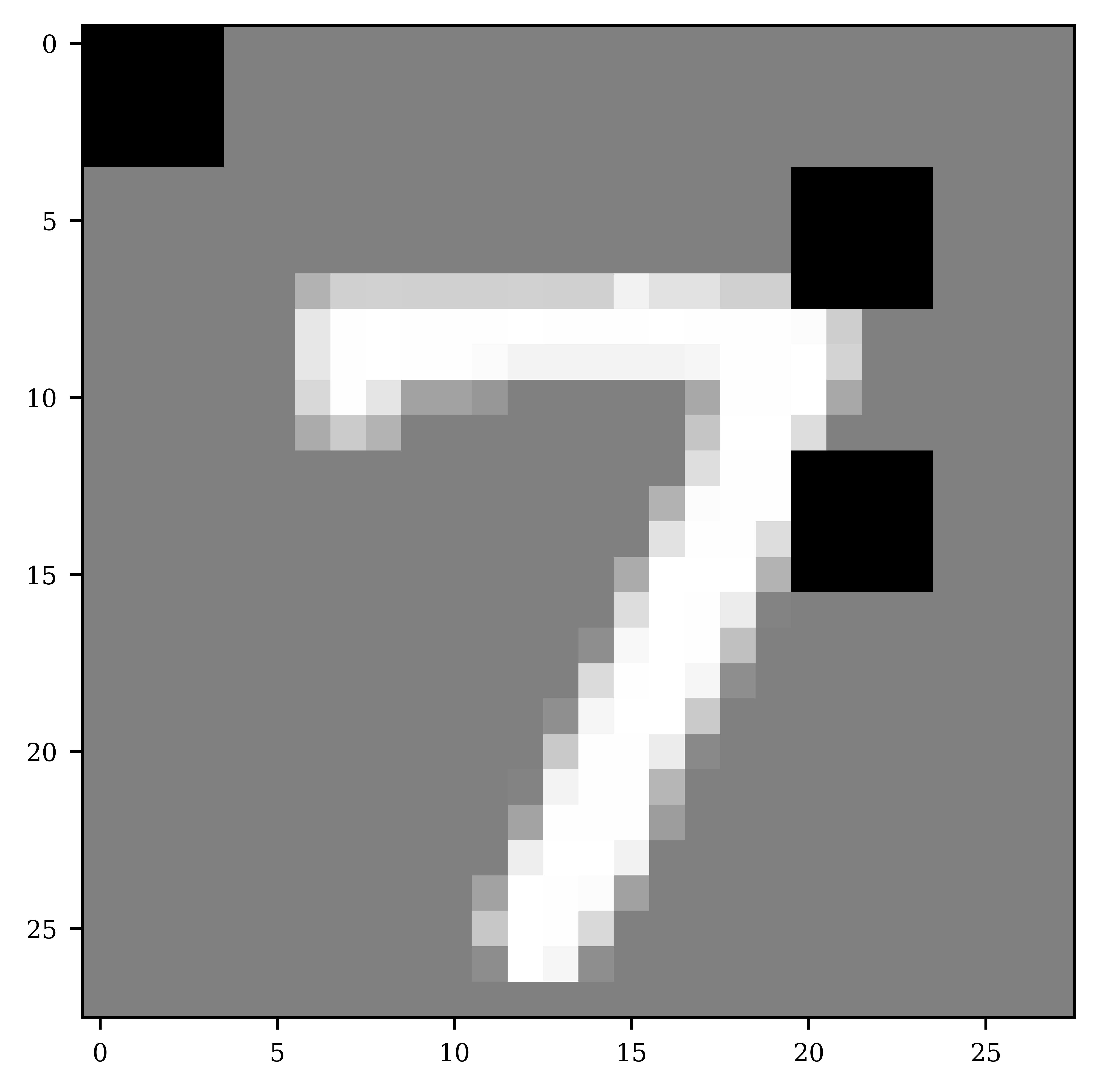}
\includegraphics[width=0.19\linewidth]{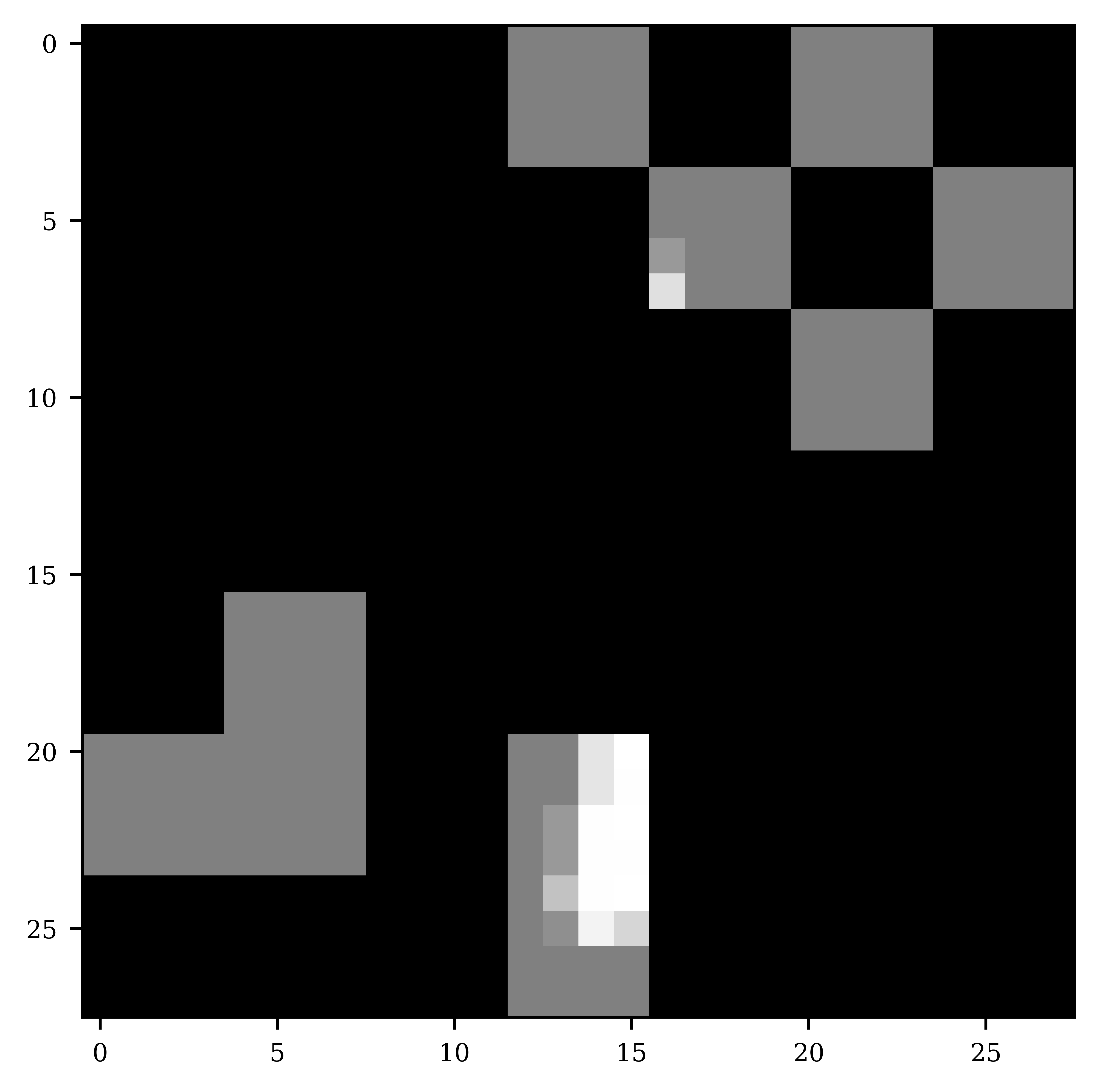}
\includegraphics[width=0.19\linewidth]{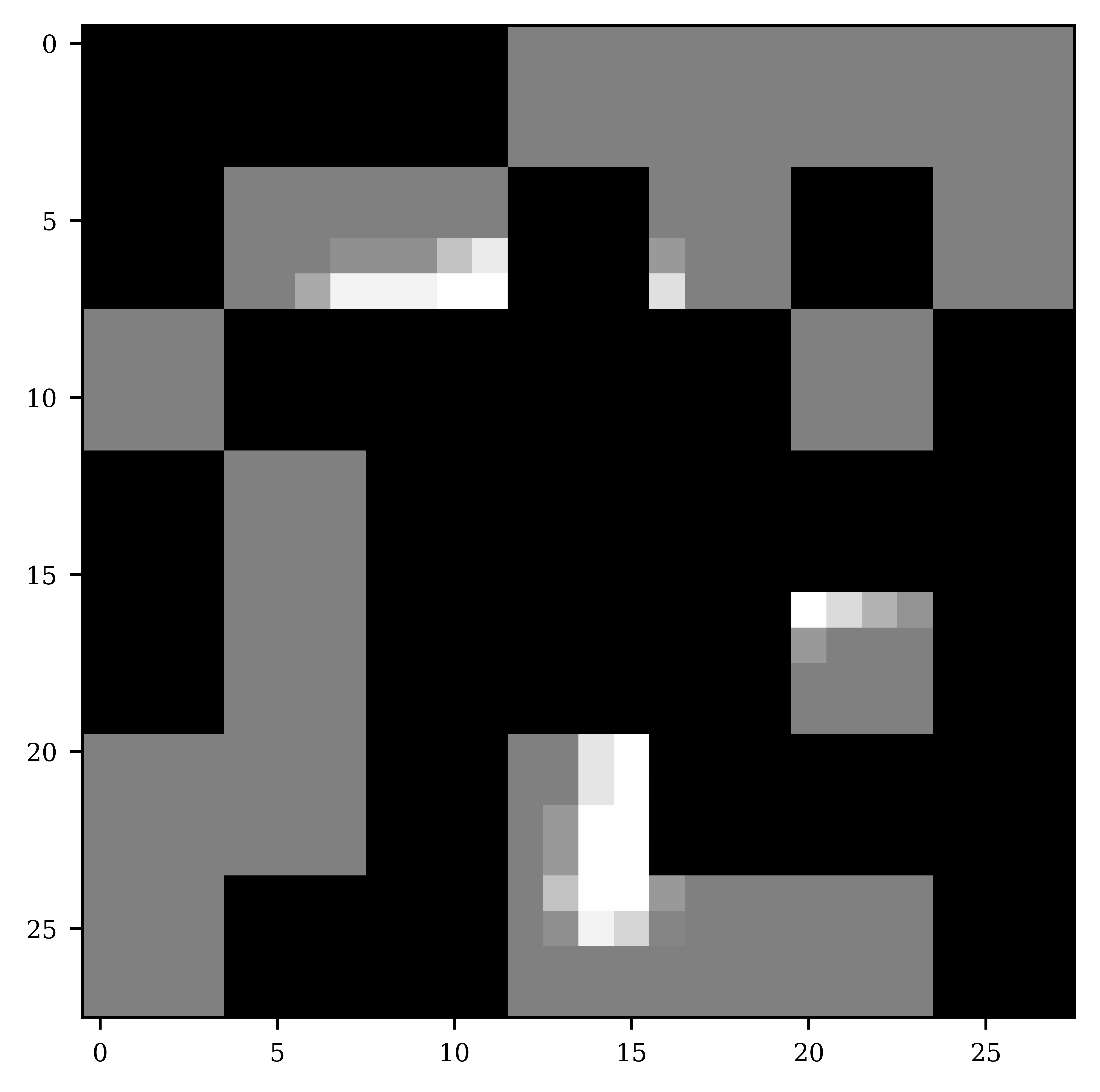}
\includegraphics[width=0.19\linewidth]{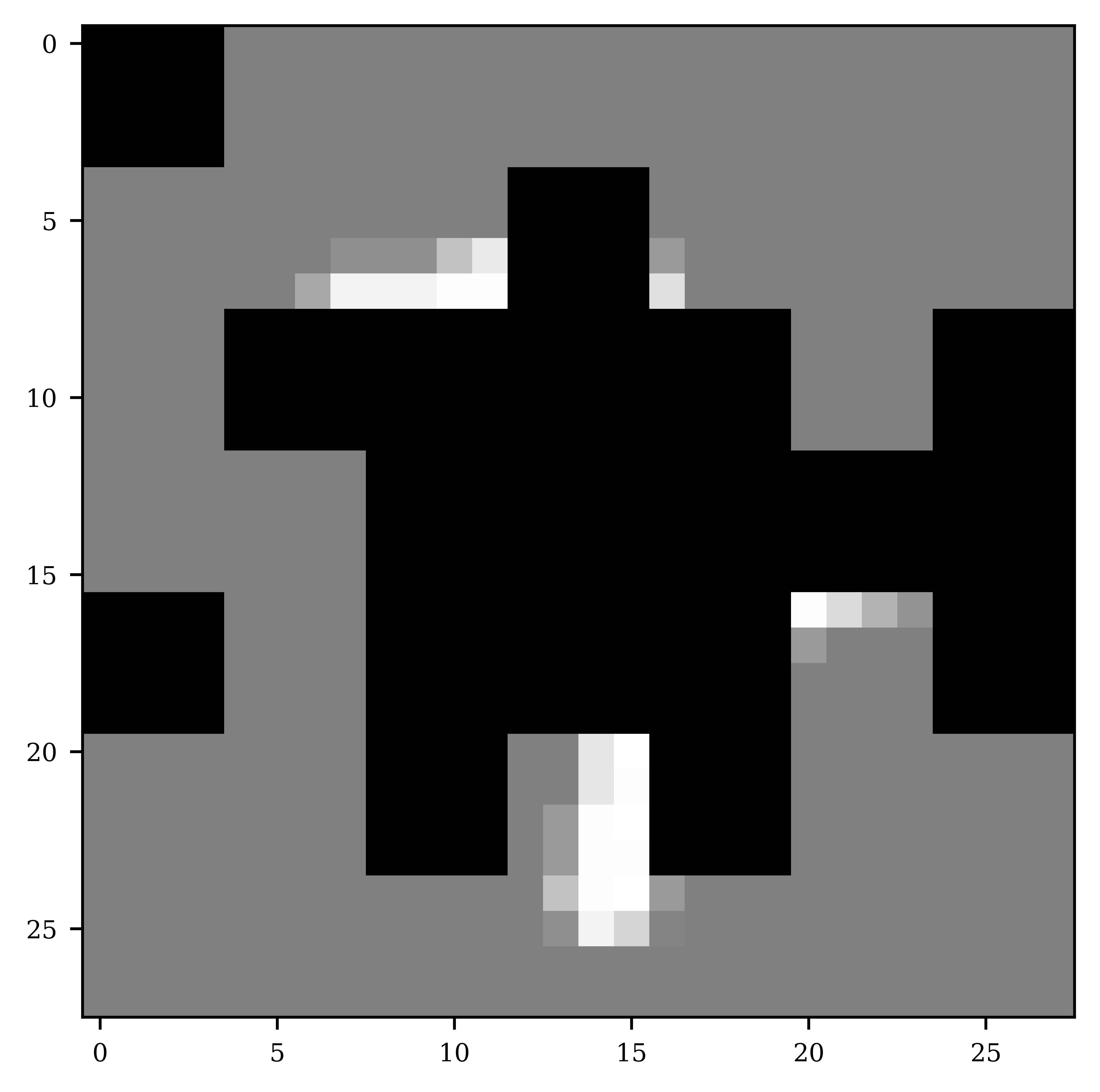}
\includegraphics[width=0.19\linewidth]{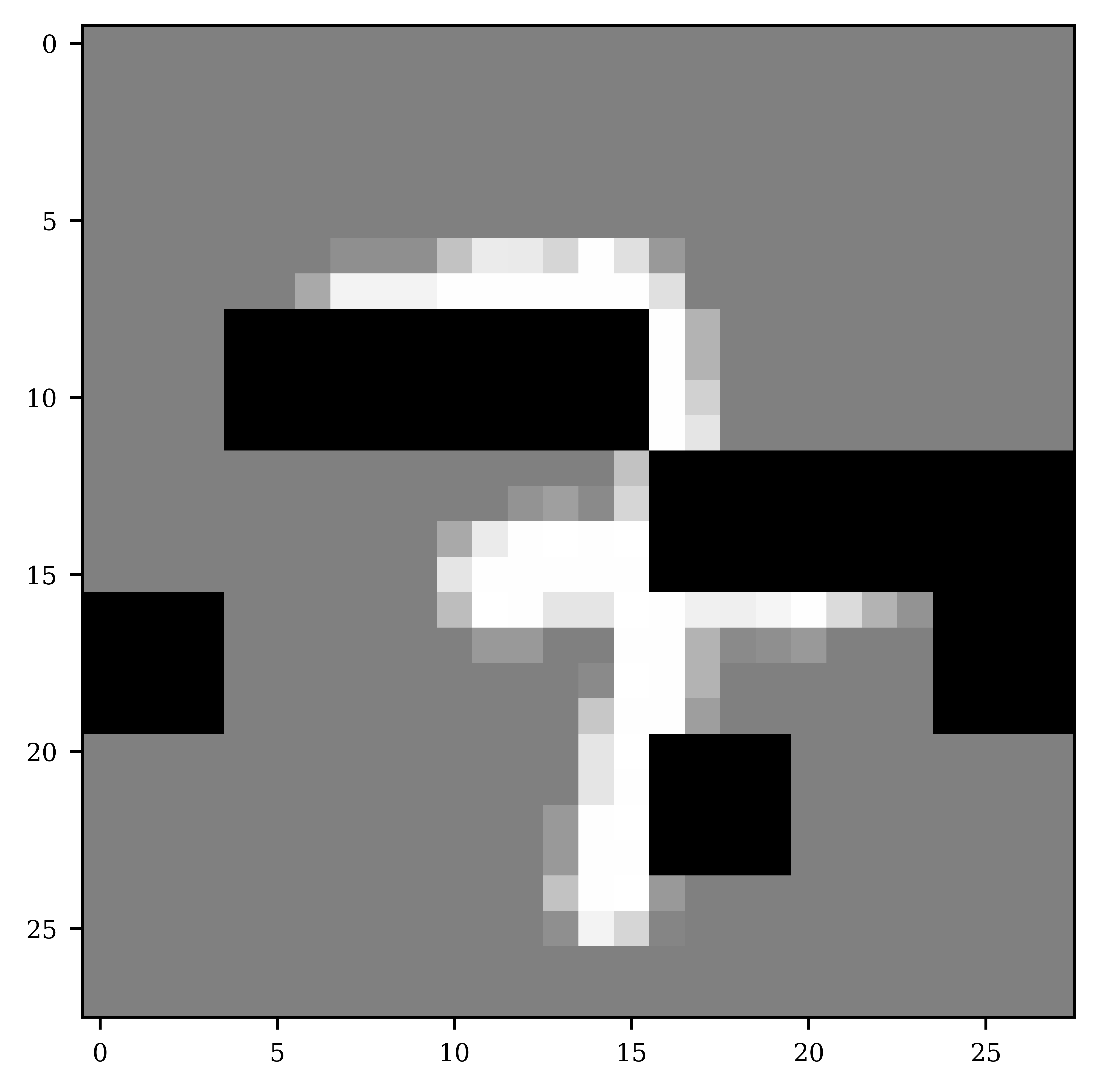}
\includegraphics[width=0.19\linewidth]{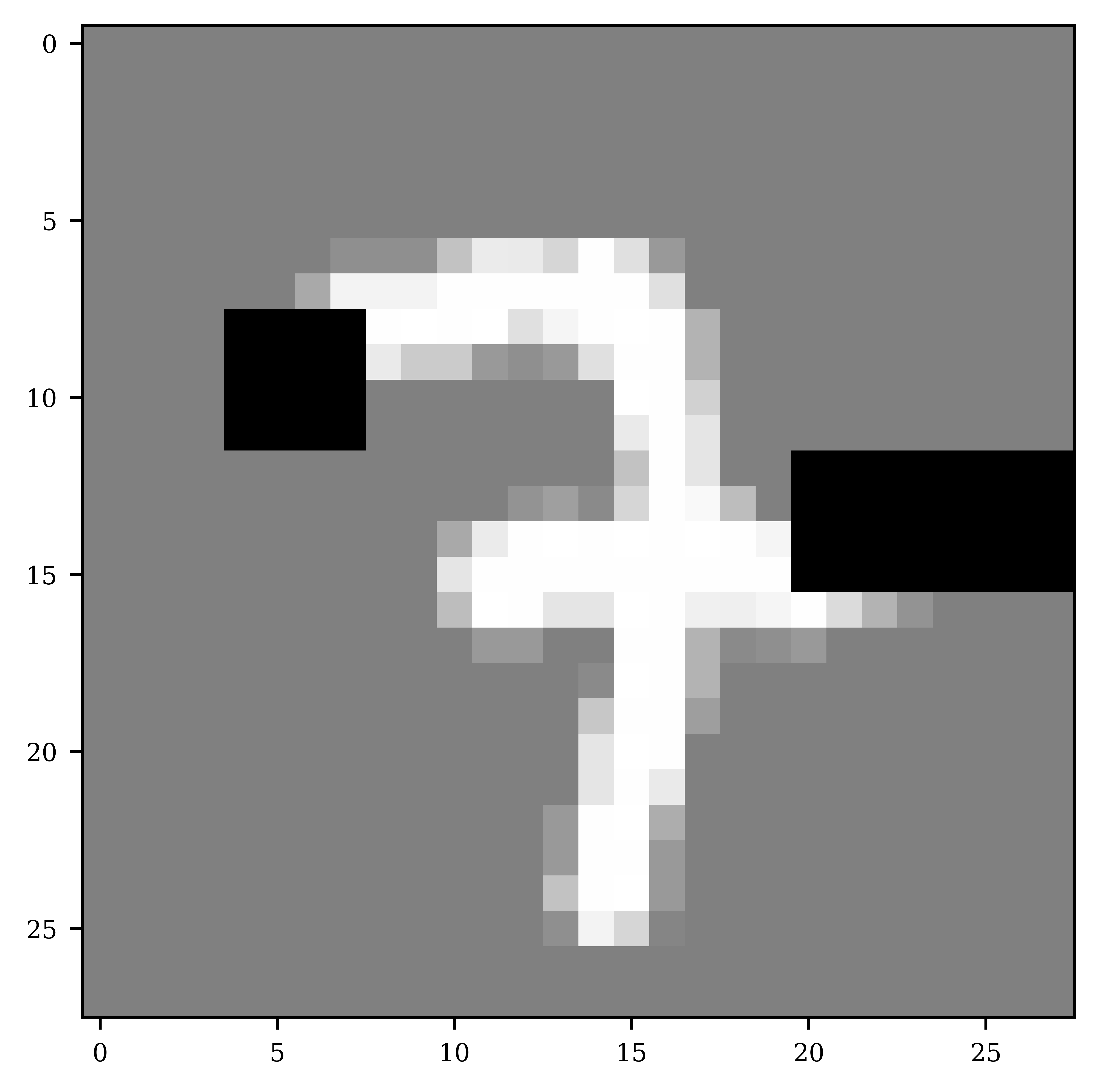}
\caption{At the number of acquired features at $10$, $20$, $30$, $40$, $46$, the unacquired features are plotted in black scale and the acquired features in gray scale. The top row shows an anticipated acquisition strategy of acquiring the informative pixels first before the background pixels. The second row exhibits a surprising acquisition strategy. The last two rows are the anticipated and surpring acquisition strategies where the cost of acquiring the features in the $16\times16$ pixel square in the middle is set to be $160$. Solutions are obtained with the SO-MCTS integrated implementation with the Random strategy for CNN.}
\label{fig:mnist}
\end{figure}

The Monte Carlo Tree Search implementations show performance improvement from the benchmark algorithms for all data sets in Figure \ref{fig:bar_chart}. Comparing the best performing MCTS implementation and the best performing benchmark algorithm, the relative improvements range from $1.2\%$ to $25.1\%$ and the logistic regression classifiers show higher improvement than the neural network classifiers with the exception of MNIST. \\
\\
\textbf{Heart Failure}:
For the logistic regression classifier (LR), the SO-MCTS integrated implementation with the Pretrain strategy is the best performer with PPO-PG with the Fit strategy as the best benchmark. The SO-MCTS standalone implementation with the Pretrain strategy performs best and PPO-AC with the Random strategy is the best benchmark for the neural network classifier.\\
\\
\textbf{Coronary Heart Disease}:
The SO-MCTS standalone implementation with the Retrain strategy is the best performer with PPO-PG with the Fit strategy as the best benchmark for LR. The MO-MCTS integrated implementation with the Random strategy performs best and PPO-PG with the Random strategy is the best benchmark for the neural network classifier.\\
\\
\textbf{PhysioNet}:
For LR, the SO-MCTS integrated implementation with the Retrain strategy is the best performer with PPO-PG with the Random strategy as the best benchmark. For the neural network classifier, the SO-MCTS integrated implementation with the Random strategy performs best and PPO-PG with the Random strategy is the best benchmark.\\
\\
\textbf{MNIST}:
The SO-MCTS integrated implementation with the Random strategy is the best performer with PPO-PG with the Random strategy as the best benchmark for LR. For the convolutional neural network classifier (CNN), the SO-MCTS integrated implementation with the Random strategy performs best and PPO-PG with the Random strategy is the best benchmark. For $10$ randomly selected samples, we also visually analyze the resulting feature acquisition sequences at the numbers of acquired features of $10$, $20$, $30$, $40$, and $46$ to determine that $70.0\%$ of the samples are acquiring the informative digit pixels first before acquiring the background pixels. In Figure \ref{fig:mnist}, the top row shows an anticipated acquisition strategy. Of the $70.0\%$ samples exhibiting the anticipated behavior, at the number of acquired features points of $10$ and $20$, the informative pixels consist of $84.0\%$ and $67.0\%$ of the acquired pixels, respectively. The second row in Figure \ref{fig:mnist} exhibits a surprising acquisition strategy. We also set the cost of acquiring the features in the $16\times16$ pixel square in the middle to be $160$ and visually compare to the case when the cost of acquiring each feature is $16$. Of the randomly selected $10$ samples, the higher cost experiment shows $25.0\%$ of the samples acquiring the informative digit pixels before the background pixels. At the number of acquired features points of $10$ and $20$, the informative pixels in this case consists of $66.0\%$ and $62.0\%$, respectively. The last two rows in Figure \ref{fig:mnist} show anticipated and surprising acquisition cases with higher cost. We note that the AUC with all equal cost is $0.556$, but with higher cost it is $0.387$ (when integrating AUCs, both maximum costs have been scaled to $1$). 
\subsubsection{Comparison of the SO and MO MCTS Implementations}
\begin{figure}
\centering 
\includegraphics[width=0.4\linewidth]{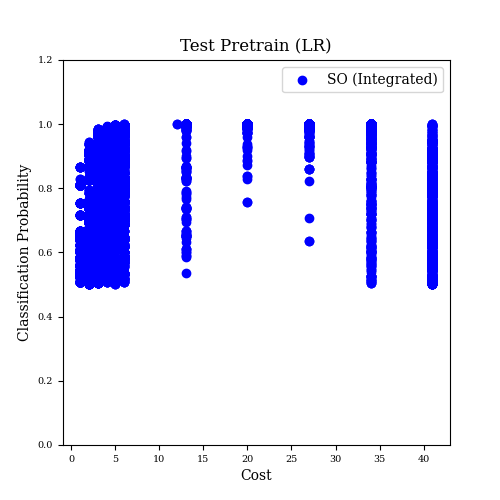}
\includegraphics[width=0.4\linewidth]{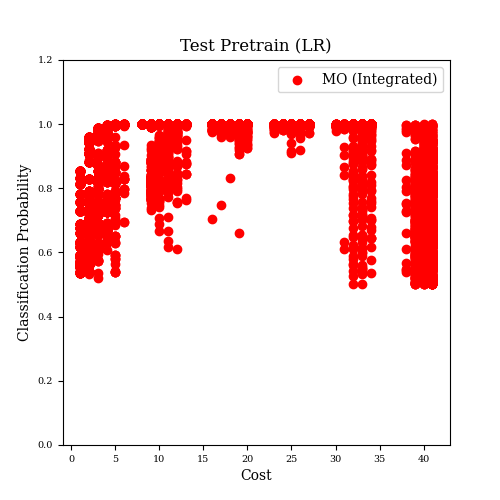}
\caption{Solutions of the SO-MCTS and MO-MCTS integrated implementations for the Heart Failure data set.}
\label{fig:hf} 
\end{figure}

\begin{figure}
\centering
\includegraphics[width=0.4\linewidth]{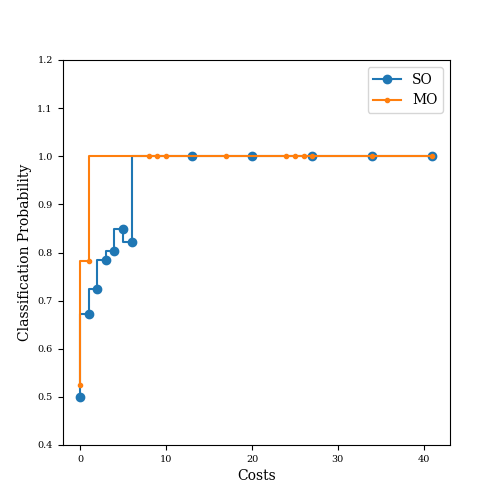}
\caption{Sample feature acquisition sequences of the SO-MCTS and MO-MCTS integrated implementations for the Heart Failure data set.}
\label{fig:trajectories}
\end{figure}

Best performance results from our MCTS implementations are shown in Table \ref{table:summary}. The results are shown as the percentages of the average F1 AUCs for each implementation with respect to the highest possible F1 AUCs of total costs of full features.
\begin{table}
\begin{center}
\begin{tabular}{l||cc||cc||cc||cc}
    \toprule
    \multirow{2}{*}{} &
      \multicolumn{2}{c||}{HF} &
      \multicolumn{2}{c||}{CHD} &
      \multicolumn{2}{c||}{PhysioNet} &
      \multicolumn{2}{c}{MNIST} \\
    & \makecell{LR \\ Mean} & \makecell{LR \\ Max} & \makecell{LR \\ Mean} & \makecell{LR \\ Max} & \makecell{LR \\ Mean} & \makecell{LR \\ Max} & \makecell{LR \\ Mean} & \makecell{LR \\ Max} \\
    \midrule
    SO-MCTS Standalone & 52.7 & 70.7 & \textbf{52.9} & 53.9 & 51.9 & 62.0 & 56.4 & 61.4 \\
    SO-MCTS Integrated & \textbf{64.4} & 67.1 & 51.6 & 53.9 & \textbf{55.2} & 61.0 & \textbf{61.1} & 64.2\\
    MO-MCTS Integrated & 59.5 & 65.9 & 49.6 & 53.3 & 46.3 & 52.2 & 57.2 & 58.9 \\
\end{tabular}
\begin{tabular}{l||cc||cc||cc||cc}
    \toprule
    \multirow{2}{*}{} &
      \multicolumn{2}{c||}{HF} &
      \multicolumn{2}{c||}{CHD} &
      \multicolumn{2}{c||}{PhysioNet} &
      \multicolumn{2}{c}{MNIST} \\
    & \makecell{NN \\ Mean} & \makecell{NN \\ Max} & \makecell{NN \\ Mean} & \makecell{NN \\ Max} & \makecell{NN \\ Mean} & \makecell{NN \\ Max} & \makecell{CNN \\ Mean} & \makecell{CNN \\ Max} \\
    \midrule
    SO-MCTS Standalone & \textbf{61.4} & 70.0 & 59.8 & 60.2 & 52.2 & 59.1 & 62.9 & 72.4 \\
    SO-MCTS Integrated & \textbf{61.4} & 71.5 & 59.0 & 62.0 & \textbf{52.5} & 55.3 & \textbf{70.3} & 77.0\\
    MO-MCTS Integrated & 60.0 & 65.9 & \textbf{63.3} & 63.7 & 52.2 & 53.6 & \textbf{70.3} & 72.0 \\
    \bottomrule
\end{tabular}
\caption{Summary tables of the MCTS implementations. Results are the percentages of the average F1 AUCs with respect to the highest possible F1 AUCs of total costs of full features. Mean are the average and max are the maximum individual experimental run.}
\label{table:summary}
\end{center}
\end{table}
With the exception of the Coronary Heart Disease data set, the SO-MCTS integrated implementation has higher F1 AUCs than the MO-MCTS integrated implementation. We plot the solutions from the Heart Failure data set in the objective space in Figure \ref{fig:hf}. In the case of the Heart Failure data set where SO-MCTS has higher F1 AUC, we see that \textbf{(1)} for lower costs, the SO-MCTS solutions are more frequent and \textbf{(2)} for higher costs, the SO-MCTS solutions are confined to cost regions that are separated by that of continuous features. This indicates that the SO-MCTS trained policy acquires the lower cost categorical features before the higher cost continuous features, whereas the MO-MCTS trained policy does not. Thus, the trained policy of acquiring the lower cost categorical features first leads to higher F1 AUCs. For the Coronary Heart Disease with the random logistic regression classifier strategy, where the MO-MCTS integrated implementation has a higher F1 AUC, the solutions in the objective space are similar to the SO-MCTS integrated implementation with the policy acquiring the lower cost categorical features first before venturing to the higher continuous features.\\
\\
We then examine the feature acquisition sequences from the Heart Failure data set in Figure \ref{fig:trajectories}. Sample solutions further validate our conclusions on the SO-MCTS and MO-MCTS trained policies. For the SO-MCTS integrated implementation, the solution acquires the lower cost categorical features with the more gradual increase in the classification probabilities before acquiriing the higher cost continuous features. For the MO-MCTS integrated implementation, the solution optimizes the classification probability and the acquisition cost simultaneously, with the increase in the classification probability dependent on the features being acquired. \\
\\
In Figure \ref{fig:trajectories}, we also observe that the MO-MCTS solution has more acquisition cost budgets under which the classification confidence threshold of $1.0$ can be reached, as the MO-MCTS solution has $10$ cost points at which this threshold is reached and the SO-MCTS solution has $5$. Since we considered the case of infinite budgets, where we obtained the ground-truth values for all the features, it is more advantageous to use the MO-MCTS implementation in tight budget situations. The MO-MCTS trained policy shows more diversity in the solution space, whereas the SO-MCTS trained policy acquires lower cost features first before being constrained to higher cost features. Thus, the MO-MCTS implementation provides more solutions matching variable budgets and confidence thresholds.
\subsubsection{Comparsion of the Standalone and Integrated Implementations}
\begin{figure}
\centering
\includegraphics[width=0.4\linewidth]{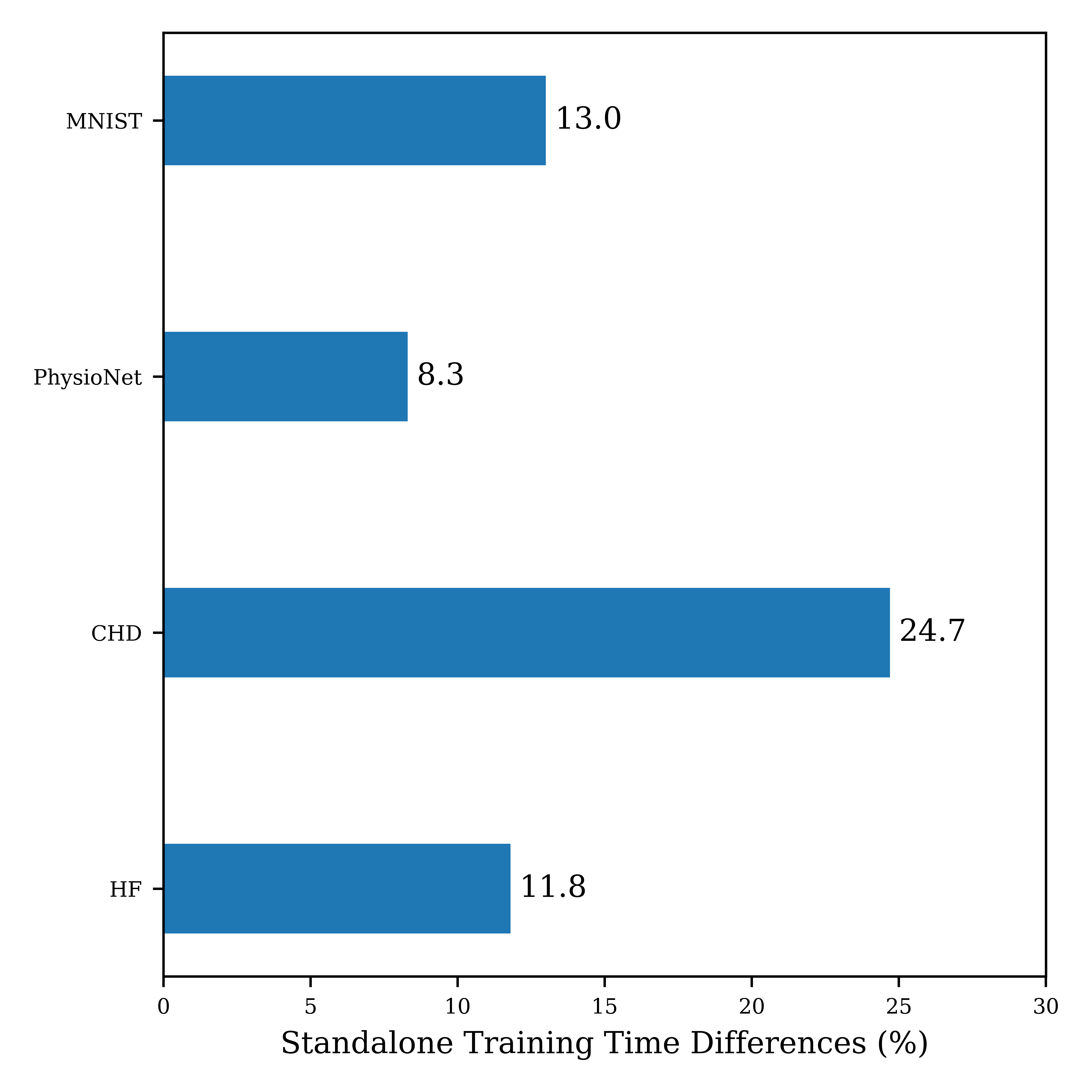}
\caption{Relative differences in the algorithm training times between the standalone and integrated implementations of the SO-MCTS.}
\label{fig:time}
\end{figure}

The F1 AUCs of the SO-MCTS integrated implementation shows relative improvement of $4.6\%$ and $1.8\%$ for the logistic regression and neural network classifiers from the SO-MCTS standalone implementation in Table \ref{table:summary}. Solutions in the objective space do not show differences between the two implementations. In Figure \ref{fig:time}, we show the relative difference in the algorithm training times for the standalone implementation from the integrated implementation, where the standalone implementation has faster relative algorithm training times by $8.3\%$ and $24.7\%$ from the integrated implementation. When the algorithm training time is another constraint in the usage of the MCTS algorithm for feature acquisition, it is advantageous to use the standalone implementation with lower training times if there is an option of slightly higher AUC. 
\subsubsection{Comparsion of the Classifier Strategies}
\begin{figure}
\centering 
\includegraphics[width=0.4\linewidth]{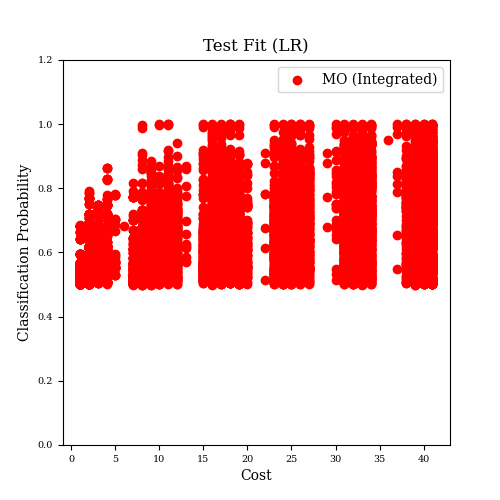}
\caption{Solutions for the MO-MCTS integrated implementation with the logistic regression classifier and fit strategy for the Heart Failure data set.}
\label{fig:fit} 
\end{figure}

For the Heart Failure, Coronary Heart Disease, and PhysioNet data sets, we use the fit strategy, where each subset of the feature set is used to train a single classifier. In comparison to the fit strategy, the best performing strategies with the SO-MCTS integrated implementation show relative performance improvements of $4.1\%$ to $24.2\%$. For the MO-MCTS integrated implementation, the best performing strategies show relative improvements of $2.4\%$ and $31.4\%$. We also plot the MO-MCTS solutions from the Heart Failure data set in the objective space in Figure \ref{fig:fit} for the logistic regression classifier with the fit strategy. In comparing the MO-MCTS solutions with the pretrain strategy in Figure \ref{fig:hf}, we observe that the solutions for the fit strategy are concentrated in the lower classification probability regions for all costs in Figure \ref{fig:fit}. Thus, in the case when we use the MO-MCTS implementation for tight budget situations, it is also advantageous to use the fit strategy, as solutions can be obtained for lower costs with slight decreases in confidence thresholds.
\subsubsection{Comparsion of the Strategies for the Unacquired Continuous Feature Values}
As described in a previous section, we also optimize a function strategy for unacquired continuous feature values in the classifiers. The optimized hyperparameters are provided in the Appendix \ref{appendix:hyperparameters}. For the logistic regression classifiers, the quadratic cost function strategy has the highest train F1 AUCs for the Heart Failure, Coronary Heart Disease and PhysioNet data sets. For the neural network classifiers, the quadratic cost function strategy has the highest train F1 AUCs for the Heart Failure data set and constant function of $0$ for the Coronary Heart Disease and PhysioNet data sets. Thus, it is advantageous to use the quadratic cost function strategies to set the values of unacquired continuous features. 
\section{Conclusions}
In this paper, we studied the feature acquisition problem, where missing features in data are acquired for ground-truth values at variable costs. To optimize the acquisition sequences, we formulated the problem as a MDP and applied our implementations of Monte Carlo Tree Search. In the single-objective implementation, the intermediary rewards for each acquisition step during the episodes are calculated based on the classification probabilities and cumulative incurred costs. In the multi-objective implementation, the classification probabilities and cumulative incurred costs are simultaneously optimized. In comparison to the Proximal Policy Optimization and Deep Q-Network algorithms, our approach shows performance improvements in all the data sets we considered, with the relative improvement in the range of $1.2\%$ to $25.1\%$. In comparing the single-objective and multi-objective implementations, the multi-objective implementation shows an advantage in budgeted situations, as it leads to more variable sequences and thus can satisfy different cost budgets and confidence thresholds. With the multi-objective implementation, the fit strategy can also be used with small budgets. The standalone implementation shows an advantage over the integrated implementation when the algorithm training time is a constraint, as it shows lower training times with slightly lower performances. For unacquired continuous features, it is advantageous to use the quadratic cost function strategies to set the values. 
\section{Acknowledgements}
We acknowledge the financial support of this research by Elevance Health, Inc. We are also grateful to Dr. Plamen Petrov for his initiation of the project.

\newpage
\appendix
\section{Pseudocodes}
\subsection{Single-objective Monte Carlo Tree Search Functions}\label{appendix:so}
\begin{algorithm}[H]
\caption{Single-objective Monte Carlo Tree Search Functions}
\begin{multicols}{2}

    \underline{function} \textbf{MCTS}($v$,$I$)\\
    \qquad \textbf{for} iteration = 1,2,$\ldots$,$I$ \textbf{do} \\
    \qquad \qquad \textbf{train}($v$) \\
    \qquad \textbf{end for} \\
    \nonl\;

    \underline{function} \textbf{train}($v$) \\
    \qquad $v_l$ = \textbf{select}($v$) \\
    \qquad \textbf{expand}($v_l$) \\
    \qquad reward = \textbf{simulate}($v_l$) \\
    \qquad \textbf{backprop}($v_l$,reward) \;
    \nonl\;

    \underline{function} \textbf{makeChild}($v$, $a$)\\
    \qquad Obtain the feature by $a$ in $s$ of $v$ to set $s'$\\
    \qquad Create node $v'$ with $s'$ where $a(v') = a$\\
    \qquad \textbf{return} $v'$\\
    \nonl\;

    \underline{function} \textbf{select}($v$) \\
    \qquad \textbf{while} True \textbf{do} \\
    \qquad \qquad \textbf{if} $v$ unexplored or terminal \textbf{do} \\
    \qquad \qquad \qquad \textbf{return} $v$ \\
    \qquad \qquad \textbf{end if} \\
    \qquad \qquad $v$ $\leftarrow$ $\argmax\limits_{v' \in C(v)}$ $\frac{Q(v')}{N(v')} + c\sqrt{\frac{\text{ln}N(v)}{N(v')}}$ \\
    \qquad \textbf{end while} \\

    \vfill\null
    \columnbreak

    \underline{function} \textbf{expand}($v$) \\
    \qquad \textbf{for} all unacquired actions $a \in$ $A(v)$ \textbf{do} \\
    \qquad \qquad $v'$ $\leftarrow$ $\textbf{makeChild}(v,a)$ \\
    \qquad \qquad Add $v'$ to $C(v)$ \\
    \qquad \qquad Set $a(v') = a$ \\
    \qquad \textbf{end for}\;
    \nonl\;

    \underline{function} \textbf{simulate}($v$) \\
    \qquad reward = 0 \\
    \qquad \textbf{while} $v$ not terminal \textbf{do} \\
    \qquad \qquad Choose $a$ $\in$ $A(v)$ uniformly at random \\
    \qquad \qquad $v$ $\leftarrow$ $\textbf{make child}(v,a)$ \\
    \qquad \qquad reward $\mathrel{+}=$ $Q(v)$ \\
    \qquad \textbf{end while}\;
    \qquad \textbf{return} reward\;
    \nonl\;

    \underline{function} \textbf{backprop}($v$, reward) \\
    \qquad \textbf{while} $v$ not null \textbf{do} \\
    \qquad \qquad $N(v)$ $\mathrel{+}= 1$ \\
    \qquad \qquad $Q(v)$ $\mathrel{+}=$ reward \\
    \qquad \qquad $v$ $\leftarrow$ parent of $v$ \\
    \qquad \textbf{end while}

\end{multicols}
\end{algorithm}
\newpage
\subsection{Multi-objective Monte Carlo Tree Search Pseudocode}\label{appendix:mo}
\begin{algorithm}[H]
\caption{Multi-objective Monte Carlo Tree Search (Integrated)}
\SetKwInOut{Input}{Input}
\Input{Iteration number $I$, initial policy network weights $\theta$, policy network update frequency $f$}
Initialize policy network $\phi$ with $\theta$\\
Initialize list $L$ of visited nodes and their $R$ and visit counts $N$\\
Initialize list $M$ of global Pareto Front approximations $P$\\
$i \leftarrow 0$ \\
\nonl\;
\underline{function} \textbf{preprocess}($L$,$M$)\\
\qquad Make each node $v$ in $L$ to be distinct with non-dominated union for $R(v)$ and $N(v)$ for duplicates\\
\qquad $A$ = $\vv{0}$\\
\qquad $S$ = $v$ in $L$ \\
\qquad \textbf{for} $v$ in $L$ \textbf{do} \\
\qquad \qquad \textbf{for} action in $A$ \textbf{do} \\
\qquad \qquad \qquad Find child nodes of $v$ in $L$ \\
\qquad \qquad \qquad \textbf{for} node in child nodes \textbf{do} \\
\qquad \qquad \qquad \qquad $R(\text{node})$ $=$ [$R(\text{node})$,$M$]\\
\qquad \qquad \qquad \qquad $A(\text{action})$ $\leftarrow$ $A(\text{action})$ $+$ \textbf{HV}($R(\text{node})$)\\
\qquad \qquad \qquad \textbf{end for}\\
\qquad \qquad \textbf{end for}\\
\qquad Normalize $A$ with division by max($A$)\\
\qquad \textbf{return} $S$, $A$\\
\nonl\;
\begin{multicols}{2}
    \textbf{for} sample = 1,2,$\ldots$,$m$ \textbf{do} \\
    \qquad $i \leftarrow i+1$ \\
    \qquad Initialize state $s_0$ \\
    \qquad Initialize global Pareto Front approximation $P$ \\
    \qquad Create root node $v_0$ with $s_0$ \\
    \qquad \qquad $R(v_0)$: local Pareto Front approximation \\
    \qquad \qquad $N(v_0)$: visit count of $v_0$ \\
    \qquad \qquad $C(v_0)$: children of $v_0$ \\
    \qquad \qquad $a(v_0)$: action of $v_0$ \\
    \qquad \textbf{while} $v_0$ not terminal \textbf{do}\\
    \qquad \qquad \textbf{MO-MCTS}$(v_0,$I$)$\\
    \qquad \qquad $a$ $\leftarrow$ $\phi_{\theta}$($s_0$)\\
    \qquad \qquad $v_0 \leftarrow \textbf{makeChild}(v_0,a)$\\
    \qquad \textbf{end while} \\
    \qquad Append $R(v)$ and $N(v)$ to $L$\\
    \qquad $M$ $\leftarrow$ \textbf{findGlobalP}($M$,$P$)\\
    \qquad \textbf{if} $f$ $\%$ $i$ == 0 \textbf{do}\\
    \qquad \qquad $S$, $A$ $\leftarrow$ \textbf{preprocess}($L$,$M$)\\
    \qquad \qquad Train $\phi_{\theta}$ on $S$ and $A$\\
    \qquad \textbf{end if} \\
    \textbf{end for}\\
    \nonl\;

    \underline{function} \textbf{MO-MCTS}($v$,$I$)\\
    \qquad \textbf{for} iteration = 1,2,$\ldots$,$I$ \textbf{do} \\
    \qquad \qquad \textbf{train}($v$) \\
    \qquad \textbf{end for} \\
    \nonl\;

    \underline{function} \textbf{train}($v$) \\
    \qquad $v_l$ = \textbf{select}($v$) \\
    \qquad \textbf{expand}($v_l$) \\
    \qquad reward = \textbf{simulate}($v_l$) \\
    \qquad \textbf{backprop}($v_l$,reward) \;
    \vfill\null
    \columnbreak

    \underline{function} \textbf{makeChild}($v$, $a$)\\
    \qquad Obtain the feature by $a$ in $s$ of $v$ to set $s'$\\
    \qquad Create node $v'$ with $s'$ where $a(v') = a$\\
    \qquad \textbf{return} $v'$\\
    \nonl\;

    \underline{function} \textbf{HV}($R(v)$)\\
    \qquad Set reference point at $[-1.0, 0.0]$ \\
    \qquad $hv = 0$ \\
    \qquad \textbf{for} front in $R(v)$ \textbf{do} \\
    \qquad \qquad $h$ = front$[i][0]$ - reference$[0]$ \\
    \qquad \qquad $hv$ $\leftarrow$ $hv +$ (front$[i][1]$ - front$[i-1][1]$)$h$ \\
    \qquad \textbf{return} $hv$\\
    \nonl\;

    \underline{function} \textbf{select}($v$) \\
    \qquad \textbf{while} True \textbf{do} \\
    \qquad \qquad \textbf{if} $v$ unexplored or terminal \textbf{do} \\
    \qquad \qquad \qquad \textbf{return} $v$ \\
    \qquad \qquad \textbf{end if} \\
    \qquad \qquad \textbf{for} $v' \in C(v)$ \textbf{do} \\
    \qquad \qquad \qquad $R(v')$ $\leftarrow$ $\frac{R(v')}{N(v')} + c\sqrt{\frac{2\text{ln}N(v)}{N(v')}}$ \\
    \qquad \qquad $v$ $\leftarrow$ $\argmax\limits_{v' \in C(v)}$ $HV(R(v'))$ \\
    \qquad \textbf{end while} \\
    \nonl\;

    \underline{function} \textbf{backprop}($v$, reward) \\
    \qquad \textbf{while} $v$ not null \textbf{do} \\
    \qquad \qquad $N(v)$ $\leftarrow N(v) + 1$ \\
    \qquad \qquad $R(v)[0]$ $\leftarrow$ $R(v)[0] + $reward$[0]$ \\
    \qquad \qquad $R(v)[1]$ $\leftarrow$ $R(v)[1] + $reward$[1]$ \\
    \qquad \qquad $P \leftarrow \textbf{findGlobalP}($P$,R(v'))$\\
    \qquad \qquad $v$ $\leftarrow$ parent of $v$ \\
    \qquad \textbf{end while}

\end{multicols}
\end{algorithm}
\newpage
\section{Experimental Setup}
\subsection{Network Architectures}\label{appendix:network}
The same network architectures are used for the neural network and convolutional neural network classifiers and policy and value networks in the algorithms, Table \ref{table:policy} and \ref{table:cnn}. 
\begin{table}[H]
\centering
\begin{tabular}{||c c c c||} 
 \hline
 Hyperparameter & Heart Failure & Coronary Heart Disease & PhysioNet \\ [0.5ex] 
 \hline\hline
 Feedforward$1$ Units & $32$ & $512$ & $256$ \\ 
 Activation & ReLU & ReLU & ReLU \\
 Feedforward$2$ Units & $16$ & $256$ & $128$ \\ 
 Activation & ReLU & ReLU & ReLU \\
 Feedforward$3$ Units & $8$ & $128$ & $64$ \\ 
 Activation & ReLU & ReLU & ReLU \\
 \hline
\end{tabular}
\caption{Neural network architectures for classification and policy and value networks in the algorithms.}
\label{table:policy}
\end{table}
\begin{table}[H]
\centering
\begin{tabular}{||c c c||} 
 \hline
 Layer & Hyperparameter & Value \\ [0.5ex] 
 \hline\hline
 Conv$1$ & \makecell{Filters \\ Kernel \\ Dilation} & \makecell{$64$ \\ $3$ \\ $2$}\\
 Activation & $-$ & ReLU \\
 Max Pooling & Pool & $2$ \\ 
 Conv$2$ & \makecell{Filters \\ Kernel \\ Dilation} & \makecell{$128$ \\ $3$ \\ $2$}\\
 Activation & $-$ & ReLU \\
 Max Pooling & Pool & $2$ \\ 
 Conv$3$ & \makecell{Filters \\ Kernel \\ Dilation} & \makecell{$256$ \\ $3$ \\ $2$}\\
 Activation & $-$ & ReLU \\
 Max Pooling & Pool & $2$ \\ 
 Final Layer & Units & $512$\\ [1ex] 
 \hline
\end{tabular}
\caption{MNIST convolutional neural network architecture for classification and feature acquisition policy.}
\label{table:cnn}
\end{table}
\subsection{Hyperparameters}\label{appendix:hyperparameters}
\subsubsection{Continuous Unacquired Feature Values}
We fitted four functions with quadratic maximum at $0$ cost, quadratic minimum at full cost, linear, and constant. The choices are shown in Tables \ref{table:hf}-\ref{table:P}.
\begin{table}[H]
    \centering
    \begin{tabular}{| c | c | c | c | c | c |}
    \hline
      Algorithms & Unacquired Features (LR) & Unacquired Features (NN) \\  \hline
    \hline
     MO-MCTS Integrated & Quad Min at 41 with $-70$ & Quad Min at 41 with $-70$ \\ \hline
     SO-MCTS Integrated & Quad Min at 41 with $-70$ & Quad Min at 41 with $-70$ \\ \hline
     SO-MCTS Integrated & Quad Max at 0 with $-50$ & Quad Min at 41 with $-50$ \\ \hline
     DQN & Quad Min at 41 with $-50$ & Quad Min at 41 with $-70$ \\ \hline
     PPO-PG & Quad Max at 0 with $-70$ & Quad Min at 41 with $-70$ \\ \hline
     PPO-AC & Quad Max at 0 with $-90$ & Quad Min at 41 with $-70$ \\ \hline
    \hline
    \end{tabular}
    \caption{Heart Failure data set.}
    \label{table:hf}
\end{table}
\begin{table}[H]
    \centering
    \begin{tabular}{| c | c | c | c | c | c |}
    \hline
      Algorithms & Unacquired Features (LR) & Unacquired Features (NN) \\  \hline
    \hline
     MO-MCTS Integrated & Quad Max at 0 with $-50$ & $0$ \\ \hline
     SO-MCTS Integrated & Quad Min at 51 with $-70$ & $0$ \\ \hline
     SO-MCTS Integrated & Quad Min at 51 with $-70$ & $0$ \\ \hline
     DQN & Quad Max at 0 with $-10$ & $0$ \\ \hline
     PPO-PG & Quad Min at 51 with $-20$ & $0$ \\ \hline
     PPO-AC & Quad Max at 0 with $-90$ & $0$ \\ \hline
    \hline
    \end{tabular}
    \caption{Coronary Heart Disease data set.}
    \label{table:chd}
\end{table}
\begin{table}[H]
    \centering
    \begin{tabular}{| c | c | c | c | c | c |}
    \hline
      Algorithms & Unacquired Features (LR) & Unacquired Features (NN) \\  \hline
    \hline
     MO-MCTS Integrated & Quad Max at 0 with $-50$ & $0$ \\ \hline
     SO-MCTS Integrated & Quad Min at 229 with $-70$ & $0$ \\ \hline
     SO-MCTS Integrated & Quad Min at 229 with $-70$ & $0$ \\ \hline
     DQN & Quad Min at 229 with $-60$ & $0$ \\ \hline
     PPO-PG & Quad Min at 229 with $-60$ & $0$ \\ \hline
     PPO-AC & Quad Min at 229 with $-60$ & $0$ \\ \hline
    \hline
    \end{tabular}
    \caption{PhysioNet data set.}
    \label{table:P}
\end{table}
\subsubsection{Hyperparameters of Algorithms}\label{appendix:algorithms_hyperparameters}
\begin{table}[H]
\centering
\begin{tabular}{||c c c c c||} 
 \hline
 Hyperparameter & Heart Failure & Coronary Heart Disease & PhysioNet & MNIST \\ [0.5ex] 
 \hline\hline
 Number of simulations & $100$ & $100$ & $100$ & $100$\\ 
 $c$ & $1.0$ & $1.0$ & $1.0$ & $1.0$ \\
 Update frequency & $18$ & $20$ & $36$ & $100$\\
 Optimizer & Adam & Adam & Adam & Adam\\
 Learning rate & $10^{-5}$ & $10^{-5}$ & $10^{-5}$ & $10^{-5}$\\ 
 Retrain frequency & $54$ & $180$ & $324$ & $10000$ \\ 
 \hline
\end{tabular}
\caption{SO-MCTS hyperparameters.}
\end{table}
\begin{table}[H]
\centering
\begin{tabular}{||c c c c c||} 
 \hline
 Hyperparameter & Heart Failure & Coronary Heart Disease & PhysioNet & MNIST \\ [0.5ex] 
 \hline\hline
 Number of simulations & $100$ & $100$ & $100$ & $100$\\ 
 $c$ & $2.0$ & $1.0$ & $1.0$ & $1.0$ \\
 Update frequency & $18$ & $20$ & $36$ & $100$\\
 Optimizer & Adam & Adam & Adam & Adam\\
 Learning rate & $10^{-5}$ & $10^{-5}$ & $10^{-5}$ & $10^{-5}$\\ 
 Retrain frequency & $18$ & $20$ & $36$ & $100$\\ 
 \hline
\end{tabular}
\caption{MO-MCTS hyperparameters.}
\end{table}
\begin{table}[H]
\centering
\begin{tabular}{||c c c c c||}
 \hline
 Hyperparameter & Heart Failure & Coronary Heart Disease & PhysioNet & MNIST \\ [0.5ex] 
 \hline\hline
 Episodes & $100$ & $100$ & $100$ & $100$\\ 
 Batch size & $6$ & $20$ & $18$ & $25$\\ 
 Update frequency & $6$ & $20$ & $18$ & $25$ \\
 $\gamma$ & $0.5$ & $0.99$ & $0.999$ & $0.99$ \\ 
 $\epsilon$-decay & $0.99$ & $0.99$ & $0.99$ & $0.5$\\
 Learning rate & $10^{-6}$ & $10^{-6}$ & $10^{-6}$ & $10^{-7}$\\ 
 Optimizer & Adam & Adam & Adam & Adam\\
 Retrain frequency & $108$ & $360$ & $684$ & $12000$\\
 \hline
\end{tabular}
\caption{DQN hyperparameters.}
\end{table}
\begin{table}[H]
\centering
\begin{tabular}{||c c c c c||}
 \hline
 Hyperparameter & Heart Failure & Coronary Heart Disease & PhysioNet & MNIST \\ [0.5ex] 
 \hline\hline
 Episodes & $100$ & $100$ & $100$ & $100$\\ 
 Clip parameter & $0.2$ & $0.2$ & $0.2$ & $0.2$\\
 GAE parameter & $0.95$ & $0.95$ & $0.95$ & $0.95$\\
 Entropy coefficient & $0.01$ & $0.01$ & $0.02$ & $0.02$\\
 Value function coefficient & $1.0$ & $1.0$ & $1.0$ & $1.0$\\
 Learning rate & $10^{-5}$ & $10^{-5}$ & $10^{-5}$ & $10^{-5}$\\ 
 Optimizer & Adam & Adam & Adam & Adam\\
 Retrain frequency & $120$ & $400$ & $360$ & $10000$\\
 \hline
\end{tabular}
\caption{PPO hyperparameters.}
\end{table}
\end{document}